%% file: ms.tex
\documentclass[10pt]{article} 
\usepackage[accepted]{tmlr}

\input{math_commands.tex}

\usepackage{comment}
\usepackage{amsmath,amssymb} 
\usepackage{color}
\usepackage[letterspace=-15]{microtype}

\usepackage[utf8]{inputenc} 
\usepackage[T1]{fontenc}    
\usepackage{hyperref}       
\usepackage{booktabs}       
\usepackage{amsfonts}       
\usepackage{nicefrac}       
\usepackage{microtype}      
\usepackage{xcolor}         
\usepackage{wrapfig}
\usepackage{csquotes}

\usepackage{graphicx}
\usepackage{amsmath}
\usepackage{bbding}
\usepackage{tabularx}
\usepackage{booktabs}
\usepackage{multirow}
\usepackage{newtxtt}
\usepackage{etoolbox,siunitx}
\robustify\bfseries
\usepackage[linesnumbered,ruled]{algorithm2e}

\usepackage[capitalize]{cleveref}
\crefname{section}{Sec.}{Section}

\usepackage[font=footnotesize,skip=2pt,subrefformat=parens]{subcaption}

\usepackage[symbol]{footmisc}

\definecolor{lightgray}{RGB}{128,128,128} 
\definecolor{lightblue}{RGB}{173,216,230}
\definecolor{vlgray}{RGB}{238, 238, 238}
\definecolor{blue}{RGB}{0,0,255} 
\definecolor{mypurple}{RGB}{255,0,255}
\definecolor{vlblue}{RGB}{217, 240, 252}

\usepackage{pifont} 

\usepackage{url}

\usepackage{xspace}
\newcommand{\eg}{\emph{e.\thinspace{}g.}\@\xspace}
\newcommand{\ie}{\emph{i.\thinspace{}e.}\@\xspace}
\newcommand{\etal}{\emph{et al.}\@\xspace}
\newcommand{\iid}{\emph{i.\thinspace{}i.\thinspace{}d.}\@\xspace}
\newcommand{\wrt}{\emph{w.\thinspace{}r.\thinspace{}t.}\@\xspace}
\newcommand{\cf}{\emph{cf.}\@\xspace}

\newcommand{\tta}{TTA\xspace}
\newcommand{\iabn}{SaN\xspace}
\newcommand{\bdd}{BDD\xspace}

\newcommand{\idd}{IDD\xspace}
\newcommand{\cityscapes}{Cityscapes\xspace}
\newcommand{\gta}{GTA\xspace}
\newcommand{\synthia}{SYNTHIA\xspace}
\newcommand{\wilddash}{WildDash\xspace}
\newcommand{\tbn}{\textit{t-}BN\xspace}
\newcommand{\pbn}{\textit{p-}BN\xspace}

\newcommand*{\myparagraph}[1]{\noindent\textbf{#1}\hspace{0.5em}}
\newcommand*{\myparagraphnospace}[1]{\noindent\textbf{#1}\hspace{0.5em}}

\newcommand{\rebuttal}[1]{\textcolor{black}{#1}}

\title{Semantic Self-adaptation: \\ Enhancing Generalization with a Single Sample}

\author{Sherwin Bahmani$^{*1}$
  \space\space
  Oliver Hahn$^{*1}$
  \space\space
  Eduard Zamfir$^{*\dagger2}$\\
  \space\space
  Nikita Araslanov$^{\dagger3}$
  \space\space
  Daniel Cremers$^{3}$
  \space\space
  Stefan Roth$^{1,4}$\\
  \textnormal{$^{1}$TU Darmstadt \space\space$^{2}$University of W\"urzburg\space\space $^{3}$TU Munich\space\space $^{4}$hessian.AI}
}

\def\month{07}  
\def\year{2023} 
\def\openreview{\url{https://openreview.net/forum?id=ILNqQhGbLx}} 

\begin{document}
\pdfoutput=1

\maketitle

\begin{abstract}
\input{sections/abstract}
\end{abstract}

\section{Introduction}
\label{sec:intro}
\input{sections/intro}
\section{Related work}
\input{sections/related}

\section{Self-adaptive learning from a single sample}
\label{sec:sada}
\input{sections/ttt}

\subsection{Self-adaptive normalization}
\label{sec:iabn}
\input{sections/iabn}

\section{Designing a principled evaluation}
\label{sec:protocol}
\input{sections/protocol}

\section{Experiments}
\label{sec:exp}

\input{sections/experiments}

\section{Conclusion}
\label{sec:conclusion}
\input{sections/conclusion}

\section*{Acknowledgments}
This project is partially funded by the European Research Council (ERC) under the European Union’s Horizon 2020 research and innovation programme (grant agreement No. 866008) as well as  the State of Hesse (Germany) through the cluster project ``The Adaptive Mind (TAM)''.

\bibliography{egbib, supp_egbib}
\bibliographystyle{tmlr}

\appendix
\newpage
\section{Baseline: Best practice implementation}
\label{sec:supp_base}
\input{supp/sections/supp_baseline}

\section{Self-adaptation: Additional analysis}
\subsection{Analyzing self-adaptive normalization (\iabn)}
\label{sec:supp_iabn}
\input{supp/sections/supp_SaN}

\subsection{Further design choices and runtime analysis}
\label{sec:supp_ttt}
\input{supp/sections/supp_SA}

\subsection{Self-adaptation with state-of-the-art architectures}
\label{sec:supp_arch}
\input{supp/sections/supp_arch}

\subsection{\rebuttal{Evaluation on diverse domain shifts}}
\label{sec:supp_diversedomains}
\input{supp/sections/supp_diversedomains}


\subsection{Qualitative examples}
\label{sec:supp_qual}
\input{supp/sections/supp_qualitative}

\subsection{Failure Cases}
\label{sec:supp_failure}
\input{supp/sections/supp_failure}

\section{Dataset details}
\label{sec:supp_data}
\input{supp/sections/supp_dataset}

\end{document}

%% file: math_commands.tex
\usepackage{amsmath,amsfonts,bm}

\def\eqref#1{equation~\ref{#1}}

\def\1{\bm{1}}

\DeclareMathAlphabet{\mathsfit}{\encodingdefault}{\sfdefault}{m}{sl}
\SetMathAlphabet{\mathsfit}{bold}{\encodingdefault}{\sfdefault}{bx}{n}

\DeclareMathOperator*{\argmax}{arg\,max}

%% file: sections/abstract.tex
The lack of out-of-domain generalization is a critical weakness of deep networks for semantic segmentation.
Previous studies relied on the assumption of a static model, \ie, once the training process is complete, model parameters remain fixed at test time.
In this work, we challenge this premise with a \emph{self-adaptive} approach for semantic segmentation that adjusts the inference process to each input sample.
Self-adaptation operates on two levels.
First, it fine-tunes the parameters of convolutional layers to the input image using consistency regularization. 
Second, in Batch Normalization layers, self-adaptation interpolates between the training and the reference distribution derived from a single test sample.
Despite both techniques being well known in the literature, their combination sets new state-of-the-art accuracy on synthetic-to-real generalization benchmarks. Our empirical study suggests that self-adaptation may complement the established practice of model regularization at training time for improving deep network generalization to out-of-domain data. Our code and pre-trained models are available at \href{https://github.com/visinf/self-adaptive}{https://github.com/visinf/self-adaptive}.

\footnotetext{${^*}$Equal contribution; \quad $\dagger$ work primarily done while at TU Darmstadt.}

%% file: sections/intro.tex

The current state of the art for semantic segmentation \citep{Chen:2018:ECA,Long:2015:FCN} lacks direly in out-of-distribution robustness, \ie, when the training and testing distributions are different.
Numerous studies have investigated this issue, primarily focusing on image classification \citep{Arjovsky:2019:IRM,Bickel:2009:DLU,Li:2017:DBA,Torralba:2011:ULD,Volpi:2018:GUD}.
However, a recent study of existing domain generalization methods \citep{Gulrajani:2020:ISL} comes to a sobering conclusion: Empirical Risk Minimization (ERM), which makes an \iid\ assumption of the training and testing samples, is still highly competitive.
This is in stark contrast to the evident advances in the area of domain adaptation, both for image classification \citep{Ben-David:2010:TLD,Ganin:2016:DAT,Long:2016:UDA,Xie:2018:SEM} and semantic segmentation \citep{Araslanov:2021:SAC_CVPR,Vu:2019:ADV,Yang:2020:FDA}.
This setup, however, assumes access to an unlabelled test distribution at training time.
In contrast, in the generalization setting considered here, \emph{only one test sample is accessible at inference time} and \emph{no knowledge between the subsequent test samples must be shared}.

In this work, we study the generalization problem of semantic segmentation from synthetic data \citep{Richter:2016:TSS,Ros:2016:SDL} through the lens of adaptation.
In contrast to previous work that modified the model architecture \citep{Pan:2018:TAO} or the training process \citep{Chen:2021:CSG,Chen:2020:ASG,Yue:2019:DRP}, we revise the standard \emph{inference} procedure with a technique inspired by domain adaptation methods \citep{Araslanov:2021:SAC_CVPR,Li:2017:RBN}.
The technique, that we term \emph{self-adaptation}, leverages a self-supervised loss, which allows for adapting to a single test sample with a few parameter updates.
Complementary to these loss-based updates, self-adaptation integrates feature statistics of the training data with those of the test sample in the Batch Normalization layers \citep{Ioffe:2015:BNA}, commonly employed in modern convolutional neural networks (CNNs) \citep{He:2016:DRL}.
Expanding upon the previous conclusions in related studies \citep{Schneider:2020:IRA}, we find that this normalization strategy not only improves the segmentation accuracy, but also the calibration quality of the prediction confidence.

In summary, our contributions are the following:
\emph{(i)} We propose a self-adaptive process for model generalization in the context of semantic segmentation, where model inference adjusts to each test sample.
\emph{(ii)} We overcome deficiencies in the experimental protocol used in previous studies with a rigorous revision.
\emph{(iii)} Implemented on top of a simple baseline, self-adaptation surpasses previous work and achieves new state-of-the-art segmentation accuracy in synthetic-to-real generalization.

%% file: sections/related.tex

Our work contributes to recent research on generalization of semantic segmentation models, and relates to studies on feature normalization \citep{Pan:2018:TAO,Schneider:2020:IRA} and online learning \citep{Sun:2020:TTT}.
While the focus in previous investigations was the training strategy \citep{Yue:2019:DRP} and model design \citep{Pan:2018:TAO}, we exclusively study the test-time inference process here.
\cite{Yue:2019:DRP} augment the synthetic training data by transferring style from real images.
Assuming access to a classification model trained on real images, \cite{Chen:2020:ASG} regularize the training on synthetic data by ensuring feature proximity of the two models via distillation, and seek layer-specific learning rates for improved generalization.
Advancing the distillation technique, \cite{Chen:2021:CSG} devise a contrastive loss that facilitates model invariance to standard image augmentations.
\cite{Pan:2018:TAO} heuristically add instance normalization (IN) layers to the network.
More recently, \cite{Choi:2021:Rob} and \cite{Huang:2021:FSD} extract domain-invariant feature statistics by either using an instance-selective whitening loss or frequency-based domain randomization. Similarly, \cite{nam2022gcisg} learn a style-invariant representation by using a causal framework for data generation.
\cite{Kundu:2021:ICCV} increase source domain diversity by augmenting single-domain data to virtually simulate a multi-source scenario. 
\cite{Tang:2021:ICCV} swap channel-wise statistics in feature normalization layers and learn adapter functions to re-adjust the mean and variance based on the input sample.
\cite{lee2022cross} enforces consistency of the output logits across multiple images (or pixels) of the same class.
To improve generalization in federated learning, \cite{caldarola2022improving} train clients locally with sharpness-aware minimization and averaging stochastic weights.
However, these methods assume access to a \emph{distribution} of real images during training \citep{Chen:2020:ASG,Chen:2021:CSG,Yue:2019:DRP} (as opposed to only for pre-training of the backbone), or require a modification of the network architecture \citep{Pan:2018:TAO}.
Our work requires neither, hence the presented technique applies even \emph{post-hoc} to the already (pre-)trained models to improve their generalization.
Moreover, as we discuss in \cref{sec:exp}, the evaluation protocol used by previous studies exhibits a number of shortcomings, which we also address.

\myparagraph{Normalization.} Batch Normalization \citep[BN;][]{Ioffe:2015:BNA} and other normalization techniques have been increasingly linked to model robustness \citep{Deecke:2019:MN,Huang:2019:INB,Schneider:2020:IRA,Wang:2018:KNN,Wu:2020:GN}.
The most commonly used BN, Layer Normalization \citep[LN;][]{Ba:2016:LAN}, and Instance Normalization \citep[IN;][]{Ulyanov:2016:INT} also affect the model's expressive power, which can be further enhanced by combining these techniques in one architecture \citep{Luo:2019:DLN,Nam:2018:BIN}.
In a domain adaptation setting, \cite{Li:2017:RBN} use source-domain statistics during training while replacing them with target-domain statistics during inference.
More recently, \cite{Schneider:2020:IRA} combine the source and target statistics during inference, but the statistics are weighted depending on the number of samples that these statistics aggregate.
\cite{Nado:2020:EPT} propose using batch statistics during inference from the target domain instead of the training statistics acquired from the source domain.
Our comprehensive empirical study complements these results by demonstrating improved generalization of semantic segmentation models.

\myparagraph{Adapting the model to a test sample.}
Several examples from previous work update the model parameters at the time of inference.
In object tracking, the object detector has the need to adjust to the changing appearance model of the tracked instance \citep{Kalal:2012:TLD}.
Conditional generative models can learn from a single image sample for the task of super-resolution \citep{Glasner:2009:SRF} and scene synthesis \citep{Shaham:2019:SLG}.
More recently, this principle has been used for improving the robustness of image classification models \citep{Sun:2020:TTT,Wang:2020:FTT,Zhang:2022:MEMO}.
The design of the self-supervised task to perform on the test sample is crucial, and the techniques developed for image classification do not always extend to dense prediction tasks, such as semantic segmentation considered here.
Nevertheless, more suitable alternatives for the self-supervised loss have been recently proposed for domain adaptation \citep{Araslanov:2021:SAC_CVPR}, and a number of other works devised domain-specific approaches for medical imaging \citep{Varsavsky:2020:TTU} or first-person vision \citep{Cai:2020:Generalizing}.
Concurrently to our work, \cite{reddy2022master} enforce edge consistency during inference for unsupervised test-time adaptation of semantic segmentation methods.

\myparagraph{Setup comparison.}
Most of these technically related works \citep{Schneider:2020:IRA,Sun:2020:TTT,Wang:2020:FTT} focus on the problem of domain adaptation in the context of image classification.
They typically assume access to a number of samples (or even all test images) from the target distribution at training time.
Our work instead addresses semantic segmentation in the domain generalization setting, which is different as it only necessitates \emph{a single datum} from the test set.
In this scenario, simple objectives, such as entropy minimization employed by Tent \citep{Wang:2020:FTT}, improve the baseline accuracy only moderately.
By contrast, our self-adaptation with pseudo-labels accounts for the inherent uncertainty in the predictions, which proves substantially more effective, as the comparison to Tent in \cref{sec:exp_iabn_ttt} reveals.
Our task is also different from few-shot learning \citep[\eg,][]{Finn:2017:MAML}, where the model may adapt at test time using a small \emph{annotated} set of image samples.
No such annotation is available in our setup; our model adjusts to the test sample in a completely unsupervised fashion, requires neither proxy tasks to update the parameters \citep{Sun:2020:TTT} nor any prior knowledge of the test distribution.

%% file: sections/ttt.tex

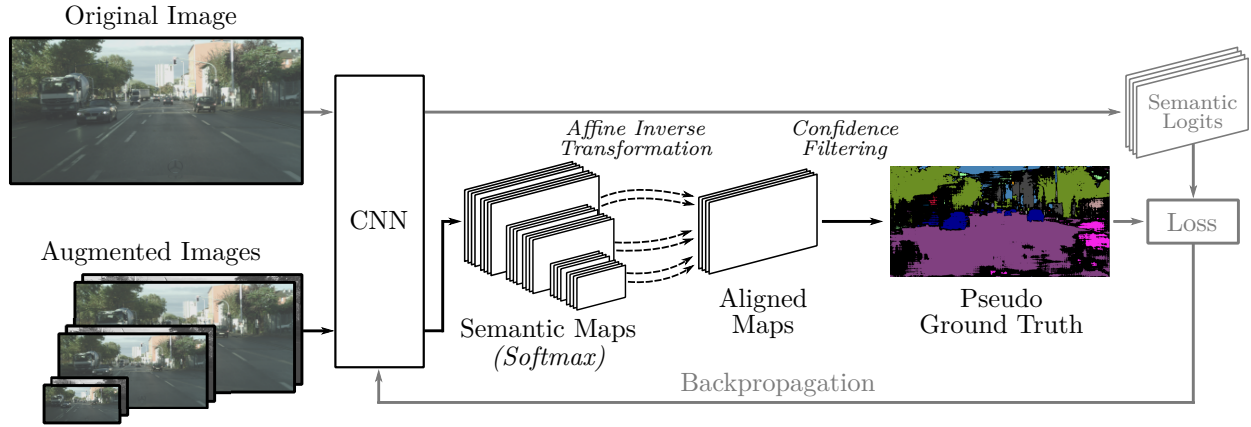
\begin{figure*}[t!]%
   \def\svgwidth{1.0\linewidth}
   \input{figures/method/tttschemenew}
   \caption{\textit{Overview of the one-sample adaptation process.} We augment a single test sample by creating a batch of images at multiple scales, each with horizontal flipping and grayscaling. To transform the output of each version back to the original image plane, we apply the corresponding inverse affine transformation to every prediction. After averaging the softmax probabilities, we create a pseudo-label using a class-dependent confidence threshold. We update the model parameters by minimizing the cross-entropy loss with respect to the pseudo-label, repeating this process for a small number of iterations ($N_t$) before producing the final prediction. The updated model is then discarded.}
   \label{fig:tttscheme}
\end{figure*}

\input{supp/sections/supp_algo}

The traditional assumption at inference time is that the parameters of the segmentation model remain fixed.
However, (self-)adaptive systems and their biological counterparts provide an example, where they learn to specialize on the particularities of their environment \citep{Thrun:1998:LLA,Widmer:1996:LPC}.
By analogy, we here allow the segmentation model to update its parameters.
Note that our setup is distinct from the domain adaptation scenario \citep[\eg, in contrast to][]{Wang:2020:FTT}, since we discard the updated parameters when processing the next sample in line with the underlying concept of domain generalization.

Our approach, visualized in \cref{fig:tttscheme} and summarized in \cref{algo:main}, uses data augmentation as a method to create mini-batches of images for each test sample.
Based on the original test image, we first create a set of $N$ augmented images by multi-scaling, horizontal flipping, and grayscaling.
These augmented images are used to form a mini-batch, which is fed through the CNN.
We transform the produced softmax probabilities from the model back to the original pixels using the inverse affine transformations, and denote the result as $m_{i,:,:,:}$ for every sample $i$ in the mini-batch.
This allows the model to have multiple predictions for one pixel.
We then compute the mean $\Bar{m}$ of these softmax probabilities along the mini-batch dimension $i$ for class $c$ and pixel $(j,k)$ on the spatial grid as
\begin{equation}
\Bar{m}_{c,j,k} = \frac{1}{N} \sum_{i=1}^N m_{i, c, j, k}.
\label{eq:mean_soft}
\end{equation}
Using hyperparameter $\psi \in (0, 1)$, we compute a threshold value $t_c$ from the maximum probability of every class to yield a class-dependent threshold $t_{c}$:
\begin{equation}
t_{c} =  \psi \cdot \max (\Bar{m}_{c,:,:}).
\label{eq:class_threshold}
\end{equation}
Finally, for every pixel, we extract the class $c^{\ast}_{j,k}$ with the highest probability by 
%
\begin{equation}
c^{\ast}_{j,k} = \argmax (\Bar{m}_{:,j,k}).
\label{eq:mean_arg}
\end{equation}
We ignore low-confidence predictions using our class-dependent threshold $t_{c}$.
Specifically, all pixels with a softmax probability below the threshold are set to an ignore label, while the remaining pixels use the dominant class $c^{\ast}_{j,k}$ as the pseudo-label $u_{j,k}$,
\begin{equation}\label{eq:pseudo_gt}
u_{j,k} =
\begin{cases}
      c^{\ast}_{j,k},& \text{if } \max (\Bar{m}_{:,j,k})\geq t_{c^{\ast}_{j,k}}\\
      \text{ignore},& \text{otherwise.}\\
   \end{cases}
\end{equation}
The pseudo ground truth $u$ for the test image is used to fine-tune the model for $N_t$ iterations with gradient descent using the cross-entropy loss.
We determine all hyperparameters, \ie, resolution of the scales, threshold $\psi$, number of iterations $N_t$, and learning rate $\eta$, based on a validation dataset.
After the self-adaptation process, we produce a single final prediction using the updated model weights.
To process the next test sample, we reset these weights to their initial value, hence the model obtains no knowledge about the complete target data distribution.

One natural consideration in this process is the quality of the confidence calibration in the semantic maps, since the $\argmax$ operation in \cref{eq:mean_arg} and applying the threshold in \cref{eq:pseudo_gt} aim to select only the most confident pixel predictions.
If the confidence values become miscalibrated (\eg, due to the domain shift), a significant fraction of incorrect pixel labels will end up in the pseudo-mask.
Our self-adaptive normalization, which we detail next, mitigates this issue.

%% file: figures/method/tttschemenew.tex
\begingroup%
  \makeatletter%
  \providecommand\color[2][]{%
    \errmessage{(Inkscape) Color is used for the text in Inkscape, but the package 'color.sty' is not loaded}%
    \renewcommand\color[2][]{}%
  }%
  \providecommand\transparent[1]{%
    \errmessage{(Inkscape) Transparency is used (non-zero) for the text in Inkscape, but the package 'transparent.sty' is not loaded}%
    \renewcommand\transparent[1]{}%
  }%
  \providecommand\rotatebox[2]{#2}%
  \newcommand*\fsize{\dimexpr\f@size pt\relax}%
  \newcommand*\lineheight[1]{\fontsize{\fsize}{#1\fsize}\selectfont}%
  \ifx\svgwidth\undefined%
    \setlength{\unitlength}{1559.05511811bp}%
    \ifx\svgscale\undefined%
      \relax%
    \else%
      \setlength{\unitlength}{\unitlength * \real{\svgscale}}%
    \fi%
  \else%
    \setlength{\unitlength}{\svgwidth}%
  \fi%
  \global\let\svgwidth\undefined%
  \global\let\svgscale\undefined%
  \makeatother%
  \begin{picture}(1,0.34)%
    \lineheight{1}%
    \setlength\tabcolsep{0pt}%
    \put(0,0){\includegraphics[width=\unitlength]{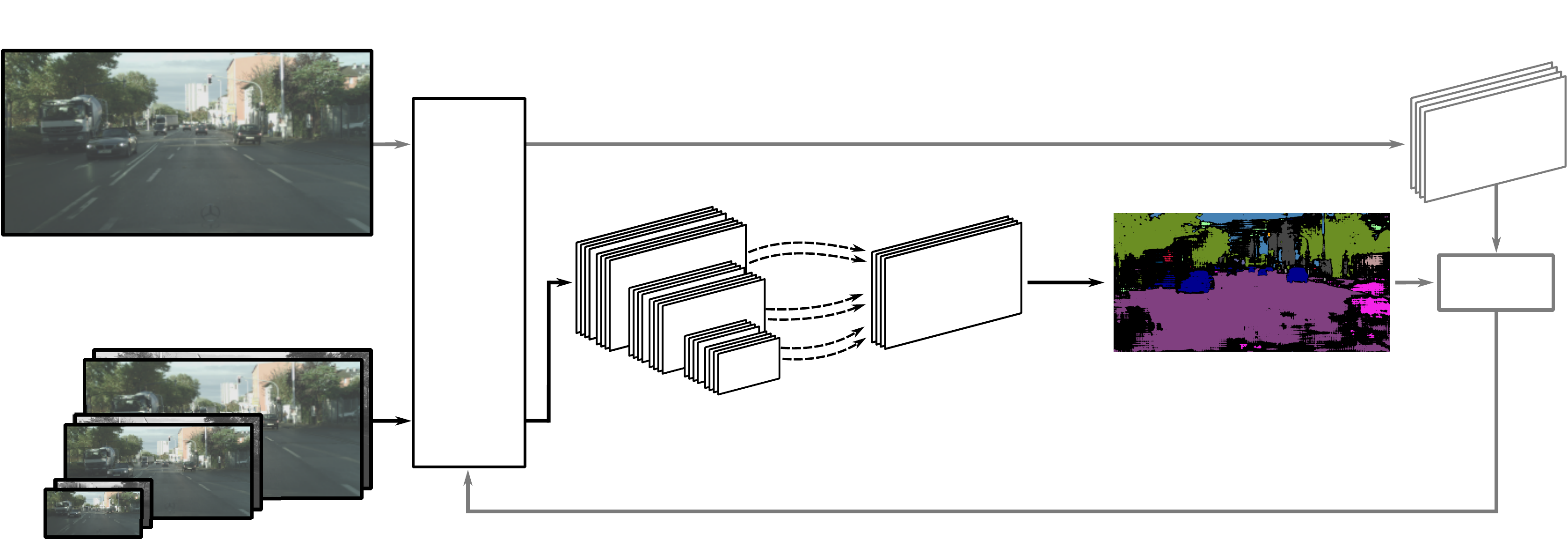}}%
    \put(0.115,0.133){\color[rgb]{0,0,0}\makebox(0,0)[ct]{\lineheight{0}\smash{\begin{tabular}[t]{l}Augmented Images \end{tabular}}}}%
    
    \put(0.115,0.323){\color[rgb]{0,0,0}\makebox(0,0)[ct]{\lineheight{0}\smash{\begin{tabular}[t]{l}Original Image \end{tabular}}}}%
    
    \put(0.275,0.158){\color[rgb]{0,0,0}\makebox(0,0)[lt]{\lineheight{1.25}\smash{\begin{tabular}[t]{l}CNN\end{tabular}}}}%
    
    \put(0.435,0.0705){\color[rgb]{0,0,0}\makebox(0,0)[ct]{\lineheight{1.0}\smash{\begin{tabular}[t]{c}Semantic Maps\\\textit{(Softmax)}\end{tabular}}}}%
    
    \put(0.608,0.095){\color[rgb]{0,0,0}\makebox(0,0)[ct]{\lineheight{1.0}\smash{\begin{tabular}[t]{c}Aligned\\Maps\end{tabular}}}}%
    
    \put(0.8,0.095){\color[rgb]{0,0,0}\makebox(0,0)[ct]{\lineheight{1.0}\smash{\begin{tabular}[t]{c}Pseudo\\Ground Truth\end{tabular}}}}%
    
    \put(0.933,0.156){\color[rgb]{0.48235294,0.48235294,0.48235294}\makebox(0,0)[lt]{\lineheight{0}\smash{\begin{tabular}[t]{l}Loss\end{tabular}}}}%
    
    
    \put(0.62,0.027){\color[rgb]{0.48235294,0.48235294,0.48235294}\makebox(0,0)[ct]{\lineheight{0}\smash{\begin{tabular}[t]{c}Backpropagation\end{tabular}}}}%
    
    \put(0.918,0.246){\color[rgb]{0.48235294,0.48235294,0.48235294}\makebox(0,0)[lt]{\lineheight{0.5}\smash{\begin{tabular}[ct]{c}\footnotesize{Semantic}\\\footnotesize{Logits}\end{tabular}}}}%
    
    \put(0.443,0.234){\color[rgb]{0,0,0}\makebox(0,0)[lt]{\lineheight{0.7}\smash{\begin{tabular}[t]{c}\textit{\footnotesize{Affine Inverse}}\\\textit{\footnotesize{Transformation}}\end{tabular}}}}%
    
    \put(0.63,0.234){\color[rgb]{0,0,0}\makebox(0,0)[lt]{\lineheight{0.7}\smash{\begin{tabular}[t]{c}\textit{\footnotesize{Confidence}}\\\textit{\footnotesize{Filtering}}\end{tabular}}}}%
  \end{picture}%
\endgroup%

%% file: supp/sections/supp_algo.tex
\begin{algorithm}
    \caption{\color{black}Summary of self-adaptation.}\label{algo:main}
     \color{black}
    
    Train segmentation model on source data (best-practice, established methodology)
    
    Replace BatchNorm with SaN (\cf \cref{sec:iabn})
    
    Tune hyperparameter $\alpha$ in SaN on validation set (WildDash)
    \\
    \# Inference on any dataset. Initial model parameters: $\theta_0$.
    
    \ForEach {$\text{test sample}$}
    {Obtain $\theta^\ast$ by minimizing cross-entropy w.r.t. pseudo-labels in \cref{eq:pseudo_gt}.\\
    Predict segmentation for the test sample using $\theta^\ast$.\\
    Reset model parameters to $\theta_0$.
    }
\end{algorithm}

%% file: sections/iabn.tex

\textbf{Batch Normalization} \citep[BN;][]{Ioffe:2015:BNA} has become an inextricable component of modern CNNs \citep{He:2016:DRL}.
Although BN was originally designed for improving training convergence, there is now substantial evidence that it plays an important role in model robustness \citep{Nado:2020:EPT}, including domain generalization \citep{Pan:2018:TAO}.
Let $x \in \mathbb{R}^{B,H,W}$ denote a spatial tensor of size $H \times W$ for an arbitrary feature channel and batch size $B$, produced by a convolutional layer.
We omit the layer and channel indexing, as the following presentation applies to all layers and feature channels on which BN operates.
At training time, BN first computes the mean and the standard deviation across the batch and spatial dimensions, \ie,
\begin{equation}
\mu = \frac{1}{B H W} \sum_{i,j,k} x_{i,j,k} \:,\qquad \sigma^2 = \frac{1}{B H W} \sum_{i,j,k} (x_{i,j,k} - \mu)^2.
\label{eq:bn_stats}
\end{equation}
The normalized features $\hat{x}$ follow from applying these statistics:
\begin{equation}
\bar{x}_{i,j,k} = \frac{(x_{i,j,k} - \mu)}{\sqrt{\sigma^2 + \epsilon}}.
\label{eq:bn_norm}
\end{equation}
Notably, this process differs from the normalization used at inference time.
At training time, every BN layer maintains a running estimate of $\mu$ and $\sigma$ across the training batches, which we denote here as $\hat{\mu}$ and $\hat{\sigma}$.
At test time, it is an established practice to normalize the feature values with $\hat{\mu}$ and $\hat{\sigma}$ in \cref{eq:bn_norm}, instead of the test-batch statistics.
We refer to this scheme as \emph{train BN} (\tbn).

\myparagraph{BN and generalization.}
In the context of out-of-distribution generalization, the running statistics $\hat{\mu}$ and $\hat{\sigma}$ derive from the source data and can be substantially different had they been computed using the target images.
This discrepancy is generally known as the \emph{covariate shift} problem.
Domain adaptation methods, which assume access to the (unlabelled) target distribution, often alleviate this issue with a technique referred to as Adaptive Batch Normalization \citep[AdaBN;][]{Li:2017:RBN}.
The key idea behind the method is to simply replace the source running statistics with those of the target.
Instead of alternating BN layers between the training and testing modes, recent work~\citep{Nado:2020:EPT} studies \emph{prediction-time BN} (\pbn), which replaces the running statistics from the training time, $\hat{\mu}$ and $\hat{\sigma}$, with the statistics $\mu$ and $\sigma$ of the current test batch (normally comprising more than one image).
Such a seemingly innocuous change is shown to benefit model robustness for image classification \citep{Nado:2020:EPT,Schneider:2020:IRA}.

In contrast to AdaBN and \pbn, which utilize either the whole target distribution or a number of samples, our study of model generalization assumes only a single target example to be available --- the one that our model receives as the input at inference time.
A viable alternative to AdaBN and \pbn is to compute the statistics per sample, which amounts to replacing BN layers with Instance Normalization (IN) layers \citep{Ulyanov:2016:INT} after model training.
However, this may cause another extreme scenario for covariate shift.
Firstly, the sample statistics may serve only as an approximation to the complete test distribution.
Secondly, such a replacement may significantly interfere with the statistics of the activations in the intermediate layers with which the network was trained.
It is, therefore, unsurprising that IN layers alone were found to only hurt the discriminative power of the model in previous work \citep{Pan:2018:TAO}.

\myparagraph{Self-adaptive normalization (\iabn).}
Let the \emph{source} mean $\hat{\mu}_{s}$ and the variance $\hat{\sigma}^2_{s}$ denote the running average of the sample statistics at training time.
If we had the \emph{target} domain knowledge expressed by the sufficient statistics $\hat{\mu}_{t}$ and $\hat{\sigma}^2_{t}$, we could use those in place of $\hat{\mu}_{s}$ and $\hat{\sigma}^2_{s}$ in the BN layers to compensate for the covariate shift.
However, at test time we only have access to the sample estimates, $\mu_{t}$ and $\sigma^2_{t}$, provided by a single datum from the target distribution:
\begin{equation}
\mu_{t} = \frac{1}{H W} \sum_{j,k} z_{j,k} \:,\qquad \sigma^2_{t} = \frac{1}{H W} \sum_{j,k} \big(z_{j,k} - \mu_{t}\big)^2,
\label{eq:test_stat}
\end{equation}
where $z \in \mathbb{R}^{H,W}$ is a spatial feature channel in a CNN of the target sample.
We define $\alpha \in [0, 1]$, which denotes the change from the source ($\alpha=0$) to a \emph{reference} ($\alpha=1$) image domain.
%
At inference time we compute the new mean and variance, $\hat{\mu}_t$ and $\hat{\sigma}_t$, as follows:
\begin{equation}
\begin{split}
\hat{\mu}_{t} := (1 - \alpha) \hat{\mu}_{s} + \alpha \mu_{t} \:,\qquad
\hat{\sigma}^2_{t} := (1 - \alpha) \hat{\sigma}^2_{s} + \alpha \sigma^2_{t}.
\end{split}
\label{eq:weighted_bn}
\end{equation}
Using \cref{eq:bn_norm}, we replace $\mu$ and $\sigma^2$ with the computed $\hat{\mu}_{t}$ and $\hat{\sigma}^2_{t}$ to normalize the features of the single test sample.
Notably, this does not affect the behavior of the BN layers at training time and applies only at test time.
Since this approach combines the inductive bias coming in the form of the running statistics from the source domain with statistics extracted from a single test instance, which is an unsupervised process, we refer to it as \emph{Self-adaptive Normalization} (\iabn).

In \cref{sec:exp_iabn}, we empirically verify that \iabn consistently boosts the segmentation accuracy in the out-of-distribution scenario.
Furthermore, we also find significant improvements of model calibration in terms of the expected calibration error \citep[ECE;][]{Naeini:2015:OWC}.

\myparagraph{Related methods.}
Setting $\alpha = 0$ in \cref{eq:weighted_bn} defaults to the established procedure, \tbn, and uses only the running statistics from training on the source domain at inference time.
Conversely, $\alpha = 1$ corresponds to Instance Normalization \citep{Ulyanov:2016:INT} or, equivalently, to the \pbn strategy \citep{Nado:2020:EPT} with a batch size of $1$.
While $\mu_{t}$ and $\sigma_{t}$ are averaged over a \emph{batch} of target images in \citep{Nado:2020:EPT}, we rely on a single test sample in this work.
Our experiments in \cref{sec:exp_iabn} expand upon previous analyses \citep{Schneider:2020:IRA} with an extensive emperical study of this normalization strategy for semantic segmentation.
Batch Instance Normalization \citep{Nam:2018:BIN} used a similar weighting approach for adaptive stylization.
However, $\alpha$ was a \emph{training} parameter, whereas we use $\alpha$ only for model selection using a validation set.

%% file: sections/protocol.tex

Previous studies \citep{Chen:2020:ASG,Chen:2021:CSG,Pan:2018:TAO,Yue:2019:DRP} on domain generalization for semantic segmentation used divergent evaluation methodologies, which exacerbates the comparison and reproducibility in follow-up research.
Encouraged by similar observations \citep{Gulrajani:2020:ISL}, we first set out and then follow a number of principles reflecting the best-practice experimentation to enable a fair and reproducible evaluation.


\emph{(A) The test set must comprise multiple domains.} --
A few previous works \citep{Chen:2020:ASG,Chen:2021:CSG,Pan:2018:TAO} used only a single target domain, Cityscapes~\citep{Cordts:2016:CDS}, for testing.
However, like other research datasets, Cityscapes~\citep{Cordts:2016:CDS} is carefully curated (\eg, the same camera hardware and country was used for image capture), hence only partially represents the visual diversity of the world.
A more comprehensive approach by \citep{Yue:2019:DRP} considered a number of target domains.
However, a separate model was selected for every target domain (based on a \emph{different} validation set).
As a result, each selected model may be biased toward the chosen target domain, hence may not be indicative of strong generalization.
This leads us to the next principle:

\emph{(B) A single model must be used for all test domains.} --
This principle follows naturally from the definition of generalization, \ie the ability of the model to reliably operate (\eg in terms of accuracy) under varying test environments.
To produce a domain-specific model is the goal of another research avenue, domain adaptation.

Many works do not clearly define the strategy of model selection, such as the used validation set.
Indeed, the use of test domains for hyperparameter tuning has not been uncommon (\eg \citep{Huang:2021:FSD}).
This is especially undesirable when studying model generalization.
Therefore, we stress that:

\emph{(C) Model selection must not use test images;} and \emph{(D) The validation set must be clearly specified.} --
These principles follow naturally from the requirements of domain generalization, with principles \textit{C} and \textit{D}, in particular, being widely accepted in machine learning research.
To our surprise, we found that \emph{no previous work on domain generalization for semantic segmentation has yet fulfilled all of these principles.}
To encourage and facilitate good evaluation practice, we therefore revise the experimental protocol used so far.
We consider a practical scenario in which a supplier prepares a model for a consumer without \textit{a-priori} knowledge on where this model may be deployed, much akin to \citet{Kundu:2021:ICCV}.
On the supplier's side, we assume access to two data distributions for model training and validation, the \emph{source data} and the \emph{validation set}.
After training the model on the source data and choosing its hyperparameters on the validation set, we assess its generalization ability on three qualitatively distinct \emph{target sets}.
The average accuracy across these sets provides an estimate of the expected model accuracy for its out-of-distribution deployment on the consumer's side.
Next, we concretize the datasets used in this study, which focus on traffic scenes for compatibility with previous work \citep{Chen:2021:CSG,Pan:2018:TAO,Yue:2019:DRP}.
Notably, the scale of our study, summarised in \cref{table:datasets}, exceeds that of previous work.

In contrast, our evaluation methodology adheres to the principles of robustness and generalization by testing a single model on multiple target domains using a single set of hyperparameters without access to images from respective target datasets. This ensures that our model can perform well across a range of domains without the need for domain-specific fine-tuning or model selection. 

\begin{table}[!t]
    \caption{\textit{State-of-the-art domain generalization methods for reported source domains and target domains.}}
    \label{table:datasets}
    \centering
    \normalsize
    \begin{tabularx}{1.0\linewidth}{@{}Xcc@{\hspace{4em}}cccc@{}}
        \toprule
        \multirow{2}{*}{Method} & \multicolumn{2}{l}{\hspace{0.4em}\textit{Source Domains}} & \multicolumn{4}{c}{\textit{Target Domains}} \\
        \cmidrule(l{0.7em}r{4.2em}){2-3} \cmidrule{4-7}
        & GTA & SYNTHIA & CS & Mapillary & BDD & IDD \\
        \midrule
        DRPC \citep{Yue:2019:DRP}       &\checkmark &\checkmark &\checkmark &\checkmark &\checkmark &- \\
        ASG \citep{Chen:2020:ASG}        &\checkmark &-  &\checkmark &- &- &- \\
        CSG \citep{Chen:2021:CSG}        &\checkmark  &- &\checkmark &- &- &- \\
        RobustNet \citep{Choi:2021:Rob}  &\checkmark &- &\checkmark &\checkmark &\checkmark &- \\
        FSDR \citep{Huang:2021:FSD}       &\checkmark &\checkmark &\checkmark &\checkmark &\checkmark &- \\
        WildNet \citep{Lee:2022:CVPR} &\checkmark & - &\checkmark &\checkmark &\checkmark & -  \\
        SAN-SAW \citep{Peng:2022:CVPR} &\checkmark &\checkmark  &\checkmark &\checkmark &\checkmark & -  \\
        PIN \citep{Kim:2022:CVPR} &\checkmark &\checkmark  &\checkmark &\checkmark &\checkmark & -  \\
        GCISG \citep{nam2022gcisg} &\checkmark & -  &\checkmark & - & - & -  \\
        XDED \citep{lee2022cross} &\checkmark & -  &\checkmark &\checkmark &\checkmark & -  \\
        \textit{This work}                &\checkmark &\checkmark &\checkmark &\checkmark &\checkmark &\checkmark \\
        \bottomrule
    \end{tabularx}
    \vspace{-0.5em}
\end{table}


%% file: sections/experiments.tex
Following our discussion above, we revise the evaluation protocol as follows.
On the supplier's side, we assume access to two data distributions for model training and validation, the \emph{source data} and the \emph{validation set}.
We assess the generalization ability of the model yielded by the validation process on three qualitatively distinct \emph{target sets}.
The average accuracy across these sets provides an estimate of the expected model accuracy for its out-of-distribution deployment on the consumer's side.
Next, we concretize the datasets used in this study, which are restricted to traffic scenes for compatibility with previous work \citep{Chen:2021:CSG,Pan:2018:TAO,Yue:2019:DRP} (see supplemental material for dataset details).

\myparagraphnospace{Source data.}
We train our model on the training split of two synthetic datasets (mutually exclusive) with low-cost ground truth annotation: \gta \citep{Richter:2016:TSS} and \synthia \citep{Ros:2016:SDL}.
Importantly, these datasets exhibit visual discrepancy (\ie, domain shift) \wrt\ the real imagery, to which our model needs to generalize.

\myparagraph{Validation set.}
For model selection and hyperparameter tuning, we use the validation set of \wilddash \citep{ZendelHMSD18}.
In our scenario, the validation set is understood to be of limited quantity, owing to its more costly annotation compared to the source data.
In contrast to the training set, however, it bears closer visual resemblance to the potential target domains.

\myparagraph{Multi-target evaluation.}
Following model selection, we evaluate the single model on three target domains comprising the validation sets from Cityscapes \citep{Cordts:2016:CDS}, \bdd \citep{Yu:2020:BDD}, and IDD \citep{Varma:2019:IDD}.
The choice of these test domains stems from a number of considerations, such as the geographic origin of the scenes (Cityscapes, \bdd, and IDD were collected in Germany, North America, and India, respectively).
Geographic distinction as well as substantial differences in data acquisition (\eg, camera properties) of these datasets bring together an assortment of challenges for the segmentation model at test time.
Since the deployment site of our model is unknown, we assume a uniform prior over the target domains as our test distribution.
Under this assumption, the average of the mean accuracy across our target domains estimates the expected model accuracy.

To compare to previous works, we also evaluate on Mapillary \citep{Neuhold:2017:MVD}.
Mapillary does not publicly disclose the geographic origins of individual samples, hence is unsuitable to identify a potential location bias acquired by the model from the training data.
This is possible in our proposed evaluation protocol, since the geographic locations from Cityscapes, \bdd, and \idd do not overlap.

\myparagraph{Implementation details.}
We implement our framework in PyTorch \citep{Paszke:2019:PYT}.
Our code and pre-trained models are publicly available.
We also discuss trivial to implement, but crucially useful training details of our baseline in-depth here and in \cref{sec:supp_base}.

Following \citep{Pan:2018:TAO}, our baseline model is DeepLabv1 \citep{Chen:2015:SIS} without CRF post-processing, but the reported results also generalize to more advanced architectures (\cf \cref{sec:supp_arch}). 
We use ResNet-50 and ResNet-101 \citep{He:2016:DRL} pre-trained on ImageNet \citep{Deng:2009:IMA} as backbone.
We minimize the cross-entropy loss with an SGD optimizer and a learning rate of 0.005, decayed polynomially with the power set to 0.9.
All models are trained on the source domains for 50 epochs with batch size, momentum, and weight decay set to 4, 0.9, and 0.0001, respectively.
For data augmentation, we compute crops of random size ($0.08$ to $1.0$) of the original image size, apply a random aspect ratio (3/4 to 4/3) to the crop, and resize the result to 512 × 512 pixels.
We also use random horizontal flipping, color jitter, random blur, and grayscaling.
We train our models with SyncBN \citep{Paszke:2019:PYT} on two NVIDIA GeForce RTX 2080 GPUs.

\subsection{\iabn improves segmentation accuracy and prediction uncertainty}
\label{sec:exp_iabn}
\begin{table}[!t]
    \caption{\emph{(a)} \textit{Segmentation accuracy using \iabn.} We report the mean IoU (\%, $\uparrow$) on three target domains (Cityscapes, BDD, IDD) across both backbones. \tbn denotes train BN \citep{Ioffe:2015:BNA}, while \pbn refers to prediction-time BN \citep{Nado:2020:EPT}.
    \emph{(b)} \textit{ECE (\%, $\downarrow$) for \iabn and MC-Dropout \citep{Gal:2016:DBA}}.
    We trained the networks on \gta and report scores for the three target domains (\cf \cref{sec:supp_iabn} for results with \synthia).}
    \label{table:iabn_calibration}
    \vspace{-1em}
    \centering
    \subcaptionbox*{}{%
        \setlength\tabcolsep{2pt}
        \begin{tabularx}{0.47\linewidth}{@{}Xcccc@{}}
        \toprule
        \multirow{2}{*}{Method} & \multicolumn{4}{@{}c}{\textbf{(a)} \textit{IoU (\%, $\uparrow$)}}\\
        \cmidrule(l){2-5}
        & CS & BDD  & IDD & Mean \\
        \midrule
        ResNet-50 & & & & \\
        \ \ w/ \tbn  & 30.95 & 28.52 & 32.78 & 30.75 \\
        \ \ w/ \pbn \ & \textbf{37.71} & 31.67 & 30.85 & 33.41 \\
        \ \ w/ \iabn (\emph{Ours}) & 37.54 & \textbf{32.79} & \textbf{34.21} & \textbf{34.85} \\
        \midrule
        ResNet-101 & & & & \\
        \ \ w/ \tbn  & 32.90 & 32.54 & 30.36 & 31.93 \\
        \ \ w/ \pbn  & 39.88 & 34.30 & 33.05 & 35.74\\
        \ \ w/ \iabn (\emph{Ours}) & \textbf{42.17} & \textbf{35.40} & \textbf{33.52} & \textbf{37.03}\\
        \bottomrule
      \end{tabularx}
        \label{table:iabn_tbn_pbn}
    }\hfill
    \subcaptionbox*{}{%
    \setlength\tabcolsep{2pt}
    \begin{tabularx}{0.47\linewidth}{@{}X@{}cccc@{}}
        \toprule
        \multirow{2}{*}{Method} & \multicolumn{4}{@{}c}{\textbf{(b)} \textit{ECE (\%, $\downarrow$)}}\\
        \cmidrule{2-5}
         & CS & BDD  & IDD & Mean \\
        \midrule
        ResNet-50 & 37.28 & 35.61 & 27.73 & 33.54 \\
        \ \ w/ \iabn (\emph{Ours}) & 30.57 & 30.94 & 26.90 & 29.47 \\
        \ \ w/ MC-Dropout & 30.29 & 29.80 & 24.17 & 28.09\\
        \ \ w/ both (\emph{Ours}) & \textbf{25.50} & \textbf{27.36} & \textbf{22.62} & \textbf{25.16} \\
        \midrule
        ResNet-101 & 35.24 & 33.74 & 27.28 & 32.09\\
        \ \ w/ \iabn (\emph{Ours}) & 26.12 & 28.89 & 23.98 & 26.36 \\
        \ \ w/ MC-Dropout & 31.30 & 29.95 & 25.15 & 28.80\\
        \ \ w/ both (\emph{Ours}) & \textbf{24.44} & \textbf{28.68} & \textbf{23.32} & \textbf{25.48} \\
        \bottomrule
    \end{tabularx}
    \label{table:calibration}
    }
\end{table}

For both source domains (\gta, \synthia) in combination with all main target domains (Cityscapes, BDD, IDD), we investigate the influence of $\alpha$ on Self-adaptive Normalization (\iabn, Eq.~\ref{eq:weighted_bn}) in terms of the IoU.
Setting an optimal $\alpha$ for every target domain is infeasible in domain generalization as the target domain during inference is unknown.
Instead, we choose the optimal $\alpha$ in steps of $0.1$ based on the IoU on the validation set of \wilddash.
For the ResNet-50 backbone, we attain the highest validation IoU for both training datasets with $\alpha = 0.1$ (see \cref{sec:supp_iabn}). 
Fixing this optimal $\alpha$, we proceed with evaluating our model on the target domains.
\cref{table:iabn_tbn_pbn} reports the segmentation accuracy with this $\alpha$ that has been optimized on the validation set.
In \cref{table:iabn_calibration}(a) we report IoU scores for both backbones on generalization from GTA to Cityscapes, BDD, and IDD and compare the accuracy of the target domains with \tbn and \pbn.
Remarkably, \iabn improves the mean IoU not only of the \tbn baseline (\eg, by 4.1\% IoU with ResNet-50), which represents an established evaluation mode, but also over the more recent \pbn~\citep{Nado:2020:EPT}.
This improvement is consistent across the board, \ie irrespective of the backbone architecture and the target domain tested.
Furthermore, we found that the calibration of our models, in terms of the expected calibration error \citep[ECE;][]{Naeini:2015:OWC}, also improves.
As shown in \cref{table:iabn_calibration}(b), not only does \iabn substantially enhance the baseline, but is even competitive with the commonly used MC-Dropout method \citep{Gal:2016:DBA}.
Rather surprisingly, \iabn exhibits a complementary effect with MC-Dropout: the calibration of the predictions improves even further when both methods are used jointly.

\begin{table}[!t]
    \caption{\textit{Mean IoU (\%) with \tta \citep{Simonyan:2015:VDC} and our self-adaptation} reported across both source domains (\gta, \synthia) and three target domains (Cityscapes, BDD, IDD).}
    \label{table:TTTvsTTA}
    \smallskip
    \centering
    \begin{tabularx}{\linewidth}{@{}Xcccc@{\hspace{2em}}cccc@{}}
        \toprule
        \multirow{2}{*}{Method}  & \multicolumn{4}{c@{\hspace{2em}}}{\textit{Source: \gta}} & \multicolumn{4}{c@{}}{\textit{Source: \synthia}} \\
        \cmidrule(l{0.8em}r{2.1em}){2-5} \cmidrule{6-9}
        & CS & BDD & IDD & Mean & CS & BDD & IDD & Mean \\
        \midrule
        ResNet-50 (w/ \iabn) & 37.54 & 32.79 & 34.21 & 34.85 & 36.14 & 26.66 & 26.37 & 29.72 \\
        \ \ \tta (w/ \iabn) & 42.56 & 37.72 & 37.98 & 39.42 & 39.67 & 32.10 & 30.46 & 34.08 \\
        \ \ Self-adaptation \emph{(Ours)} & \textbf{45.13} & \textbf{39.61} & \textbf{40.32} & \textbf{41.69} & \textbf{41.60} & \textbf{33.35} & \textbf{31.22} & \textbf{35.39} \\
        \midrule
        ResNet-101 (w/ \iabn) & 42.17 & 35.40 & 33.52 & 37.03 & 38.01 & 28.66 & 27.28 & 31.32 \\
        \ \ \tta (w/ \iabn) & 44.37 & 38.49 & 38.35 & 40.40 & 39.91 & 32.68 & 30.04 & 34.21 \\
        \ \ Self-adaptation \emph{(Ours)} & \textbf{46.99} & \textbf{40.21} & \textbf{40.56} & \textbf{42.59} & \textbf{42.32} & \textbf{33.27} & \textbf{31.40} & \textbf{35.66} \\
        \bottomrule
    \end{tabularx}
\end{table}

\subsection{Self-adaptation \textit{vs.} Test-Time Augmentation (TTA)}
\label{sec:exp_ttt}

We compare our self-adaptation to the standard non-adaptive inference, as well as test our models against Test-Time Augmentation \citep[\tta;][]{Simonyan:2015:VDC} as a stronger baseline.
\tta augments the test samples with their flipped and grayscaled version on multiple scales and averages the predictions as the final result.
For self-adaptation, we use horizontal flipping and grayscaling with factor scales of (0.25, 0.5, 0.75) \wrt\ the original image resolution.
We study the relative importance of these augmentation types in \cref{sec:supp_ttt}.
Based on the validation set WildDash, we set threshold $\psi = 0.7$, $N_t = 10$ iterations, and a learning rate $\eta = 0.05$.
We only train the layers conv4\_x, conv5\_x, and the classification head as we did not observe any benefits from updating all model parameters. Furthermore, this reduces runtime due to not backpropagating through the whole network. We investigate this choice as part of the runtime-accuracy trade-off in \cref{sec:exp_aux}.
In \cref{table:TTTvsTTA}, we show IoU scores for both source domains (\gta, \synthia) and three target domains (Cityscapes, BDD, IDD) across both backbones.
Even though \tta improves the baseline (\eg, by 3.37\% IoU with ResNet-101 using \gta), our proposed self-adaptation still outperforms it by a clear and consistent margin of $2.19\%$ IoU on average.
This observation aligns well with our reported ECE scores in \cref{table:calibration}(b) to demonstrate that self-adaptation further exploits the calibrated confidence of our predictions to yield reliable pseudo-labels for adapting the model to a particular test sample.

\begin{table}[!t]
  \caption{\textit{Mean IoU (\%,  $\uparrow$) comparison to state-of-the-art domain generalization methods} for both source domains (\gta, \synthia) as well as three target domains (Cityscapes, Mapillary, BDD). In-domain training to obtain the upper bounds uses our baseline DeepLabv1 following the same schedule as with the synthetic datasets. $(^\ddagger)$, $(^\dagger)$, $(^{\ddagger\ddagger})$ and $(^{\dagger\dagger})$ denote the use of FCN~\citep{Long:2015:FCN}, DeepLabv2~\citep{Chen:2018:DLS}, DeepLabv3~\citep{chen2017rethinking}, and DeepLabv3+~\citep{Chen:2018:ECA}, respectively.}
  \label{table:sota_cmp}
  \smallskip
  \centering
  \footnotesize
  \setlength\tabcolsep{1pt}
\renewcommand{\arraystretch}{0.9}
\begin{tabularx}{\linewidth}{@{}lXl@{\hspace{0.1mm}}ll@{\hspace{0.1mm}}ll@{\hspace{0.1mm}}l@{\hspace{0.6em}}l@{\hspace{0.1mm}}ll@{\hspace{0.1mm}}ll@{\hspace{0.1mm}}l@{}}
\toprule
& \multirow{2}{*}{Method} & \multicolumn{6}{@{}c@{\hspace{1em}}}{\textit{Backbone: ResNet-50}} & \multicolumn{6}{@{\hspace{1em}}c@{}}{\textit{Backbone: ResNet-101}} \\
\cmidrule(l{0.2em}r@{1.2em}){3-8} \cmidrule{9-14}
& & \multicolumn{2}{c}{CS} & \multicolumn{2}{c}{Mapillary} & \multicolumn{2}{c}{BDD} & \multicolumn{2}{c}{CS}  & \multicolumn{2}{c}{Mapillary} & \multicolumn{2}{c}{BDD} \\
\midrule
& In-domain Bound & 71.23 & & 58.39 & & 58.53 & & 73.84 & & 62.81 & & 61.19 & \\
\midrule

\multirow{21.5}{*}{\rotatebox[origin=c]{90}{\textit{\gta}}\hspace{2mm}} & No Adapt & \textcolor{lightgray}{32.45} & \multirow{2}{*}{$\uparrow$\textit{4.97}} & \textcolor{lightgray}{25.66} & \multirow{2}{*}{$\uparrow$\textit{8.46}}   & \textcolor{lightgray}{26.73} & \multirow{2}{*}{$\uparrow$\textit{5.41}} & \textcolor{lightgray}{33.56} & \multirow{2}{*}{$\uparrow$\textit{8.97}} & \textcolor{lightgray}{28.33} & \multirow{2}{*}{$\uparrow$\textit{9.72}} & \textcolor{lightgray}{27.76} & \multirow{2}{*}{$\uparrow$\textit{10.96}} \\
& DRPC$^\ddagger$ \citep{Yue:2019:DRP} & 37.42 & & 34.12 & & 32.14 & & 42.53 & & 38.05 & & 38.72 &  \\[3pt]
& No Adapt & \textcolor{lightgray}{25.88} & \multirow{2}{*}{$\uparrow$\textit{3.77}}  & & \multirow{2}{*}{--} & & \multirow{2}{*}{--} & \textcolor{lightgray}{29.63} & \multirow{2}{*}{$\uparrow$\textit{3.16}} & & \multirow{2}{*}{--} & & \multirow{2}{*}{--}\\
& ASG$^\dagger$ \citep{Chen:2020:ASG}  & 29.65 & & & & & & 32.79 & & & & & \\[3pt]
& No Adapt & \textcolor{lightgray}{25.88} & \multirow{2}{*}{$\uparrow$\textit{9.39}}  & & \multirow{2}{*}{--} & & \multirow{2}{*}{--} & \textcolor{lightgray}{29.63} & \multirow{2}{*}{$\uparrow$\textit{9.25}} & & \multirow{2}{*}{--} & & \multirow{2}{*}{--}\\
& CSG$^\dagger$ \citep{Chen:2021:CSG} & 35.27 & & & & & & 38.88 & & & & & \\[3pt]
& No Adapt & \textcolor{lightgray}{28.95} & \multirow{2}{*}{$\uparrow$\textit{7.63}} & \textcolor{lightgray}{28.18} & \multirow{2}{*}{$\uparrow$\textit{12.15}} & \textcolor{lightgray}{25.14} & \multirow{2}{*}{$\uparrow$\textit{10.06}} & & \multirow{2}{*}{--} & & \multirow{2}{*}{--} & & \multirow{2}{*}{--} \\
& RobustNet$^{\dagger\dagger}$\citep{Choi:2021:Rob} & 36.58 & & 40.33 & & 35.20 & & & & & & &  \\[3pt]
& No Adapt  & & \multirow{2}{*}{--} & & \multirow{2}{*}{--} & & \multirow{2}{*}{--} & \textcolor{lightgray}{33.4} & \multirow{2}{*}{$\uparrow$\textit{11.4}} & \textcolor{lightgray}{27.9} & \multirow{2}{*}{$\uparrow$\textit{15.5}} & \textcolor{lightgray}{27.3} & \multirow{2}{*}{$\uparrow$\textit{13.9}} \\
& FSDR$^\ddagger$ \citep{Huang:2021:FSD} & & & & & & & 44.8 & & 43.4 & & 41.20 & \\[3pt]
%
%
%
& No Adapt & \textcolor{lightgray}{35.16} & \multirow{2}{*}{$\uparrow$\textit{9.46}} & \textcolor{lightgray}{31.29} & \multirow{2}{*}{$\uparrow$\textit{14.77}} & \textcolor{lightgray}{29.71} & \multirow{2}{*}{$\uparrow$\textit{8.71}} & \textcolor{lightgray}{35.73} & \multirow{2}{*}{$\uparrow$\textit{10.06}} & \textcolor{lightgray}{33.42} & \multirow{2}{*}{$\uparrow$\textit{13.66}} & \textcolor{lightgray}{34.06} & \multirow{2}{*}{$\uparrow$\textit{7.67}} \\
& WildNet$^{\dagger\dagger}$ \citep{Lee:2022:CVPR} & 44.62 & & 46.09 & & 38.42 & & 45.79 & & 47.08 & & \textbf{41.73} & \\[3pt]
& No Adapt & \textcolor{lightgray}{29.32} & \multirow{2}{*}{$\uparrow$\textit{10.43}} & \textcolor{lightgray}{28.33} & \multirow{2}{*}{$\uparrow$\textit{13.53}} & \textcolor{lightgray}{25.71} & \multirow{2}{*}{$\uparrow$\textit{11.63}} & \textcolor{lightgray}{30.64} & \multirow{2}{*}{$\uparrow$\textit{14.69}} & \textcolor{lightgray}{28.65} & \multirow{2}{*}{$\uparrow$\textit{12.12}} & \textcolor{lightgray}{27.82} & \multirow{2}{*}{$\uparrow$\textit{13.36}} \\
& SAN-SAW \citep{Peng:2022:CVPR} & 39.75 & & 41.86 & & 37.34 & & 45.33 & & 40.77 & & 41.18 & \\[3pt]
& No Adapt & \textcolor{lightgray}{31.60} & \multirow{2}{*}{$\uparrow$\textit{9.40}} & \textcolor{lightgray}{29.00} & \multirow{2}{*}{$\uparrow$\textit{8.40}} & \textcolor{lightgray}{25.10} & \multirow{2}{*}{$\uparrow$\textit{9.50}} & -- &  & -- & & -- & \\
& PIN \citep{Kim:2022:CVPR} & 41.00$^{\dagger\dagger}$& & 37.40$^{\dagger\dagger}$ & & 34.60$^{\dagger\dagger}$ & & 44.90$^\dagger$  & & 39.71$^\dagger$  & & 41.31$^\dagger$  & \\[3pt]
& No Adapt & \textcolor{lightgray}{26.67} & \multirow{2}{*}{$\uparrow$\textit{12.34}}  & & \multirow{2}{*}{--} & & \multirow{2}{*}{--} & & \multirow{2}{*}{--} & & \multirow{2}{*}{--} & & \multirow{2}{*}{--}\\
& GCISG$^{\ddagger\ddagger}$ \citep{nam2022gcisg} & 39.01 & & & & & & & & & & & \\[3pt]
& No Adapt & \textcolor{lightgray}{28.90} & \multirow{2}{*}{$\uparrow$\textit{10.30}} & \textcolor{lightgray}{28.10} & \multirow{2}{*}{$\uparrow$\textit{9.00}} & \textcolor{lightgray}{25.10} & \multirow{2}{*}{$\uparrow$\textit{7.30}} & \multirow{2}{*}{--} & & \multirow{2}{*}{--} & & \multirow{2}{*}{--}\\
& XDED$^{\dagger\dagger}$ \citep{lee2022cross} & 39.20 & & 37.10 & & 32.40 & & & & & & & \\[3pt]
& No Adapt & \textcolor{lightgray}{30.95} & \multirow{2}{*}{$\uparrow$\textit{14.18}}  & \textcolor{lightgray}{34.56} & \multirow{2}{*}{$\uparrow$\textit{12.93}} & \textcolor{lightgray}{28.52} & \multirow{2}{*}{$\uparrow$\textit{11.09}} & \textcolor{lightgray}{32.90} & \multirow{2}{*}{$\uparrow$\textit{14.09}} & \textcolor{lightgray}{36.00} & \multirow{2}{*}{$\uparrow$\textit{11.49}} & \textcolor{lightgray}{32.54} & \multirow{2}{*}{$\uparrow$\textit{7.67}} \\
& Self-adaptation \emph{(Ours)} & \textbf{45.13} & & \textbf{47.49} & & \textbf{39.61} & & \textbf{46.99} & & \textbf{47.49} & & 40.21 & \\
\midrule
\multirow{9}{*}{\rotatebox[origin=c]{90}{\textit{\synthia}}} & No Adapt & \textcolor{lightgray}{28.36} & \multirow{2}{*}{$\uparrow$\textit{7.29}} & \textcolor{lightgray}{27.24} & \multirow{2}{*}{$\uparrow$\textit{5.50}}   & \textcolor{lightgray}{25.16} & \multirow{2}{*}{$\uparrow$\textit{6.37}} & \textcolor{lightgray}{29.67} & \multirow{2}{*}{$\uparrow$\textit{7.91}} & \textcolor{lightgray}{28.73} & \multirow{2}{*}{$\uparrow$\textit{5.39}}   & \textcolor{lightgray}{25.64} & \multirow{2}{*}{$\uparrow$\textit{8.70}} \\
& DRPC$^\ddagger$ \citep{Yue:2019:DRP} & 35.65 & & 32.74 & & 31.53 & & 37.58 & & 34.12 & & 34.34 &  \\[3pt]
& No Adapt  & & \multirow{2}{*}{--} & & \multirow{2}{*}{--} & & \multirow{2}{*}{--} & - &  & - &  & - &  \\
& FSDR$^\ddagger$ \citep{Huang:2021:FSD} & & & & & & & 40.8 & & 39.6 & & \textbf{37.4} & \\[3pt]
%
%
& No Adapt & \textcolor{lightgray}{23.18} & \multirow{2}{*}{$\uparrow$\textit{15.74}} & \textcolor{lightgray}{21.79} & \multirow{2}{*}{$\uparrow$\textit{12.73}} & \textcolor{lightgray}{24.50} & \multirow{2}{*}{$\uparrow$\textit{10.74}} & \textcolor{lightgray}{23.85} & \multirow{2}{*}{$\uparrow$\textit{17.02}} & \textcolor{lightgray}{21.84} & \multirow{2}{*}{$\uparrow$\textit{15.42}} & \textcolor{lightgray}{25.01} & \multirow{2}{*}{$\uparrow$\textit{10.97}} \\
& SAN-SAW \citep{Peng:2022:CVPR} & 38.92  & & 34.52 & & \textbf{35.24} & & 40.87 & & 37.26 & & 35.98 & \\[3pt]
& No Adapt & \textcolor{lightgray}{31.83} & \multirow{2}{*}{$\uparrow$\textit{9.77}} & \textcolor{lightgray}{33.41} & \multirow{2}{*}{$\uparrow$\textit{7.80}} & \textcolor{lightgray}{24.30} & \multirow{2}{*}{$\uparrow$\textit{9.05}} & \textcolor{lightgray}{37.25} & \multirow{2}{*}{$\uparrow$\textit{5.07}} & \textcolor{lightgray}{36.84} & \multirow{2}{*}{$\uparrow$\textit{4.36}} & \textcolor{lightgray}{29.32} & \multirow{2}{*}{$\uparrow$\textit{3.95}} \\
& Self-adaptation \emph{(Ours)} & \textbf{41.60} & & \textbf{41.21} & & 33.35 & & \textbf{42.32} & & \textbf{41.20} & & 33.27 & \\
\bottomrule
\end{tabularx}
\end{table}

\subsection{Comparison to state of the art}
\label{sec:exp_iabn_ttt}

We compare self-adaptation with state-of-the-art domain generalization methods in \cref{table:sota_cmp}.
While most of the other methods report their results on weakly tuned baselines, we show consistent improvements even over a carefully tuned baseline, regardless of backbone architecture or source data.
Our single model with self-adaptation even outperforms DRPC \cite{Yue:2019:DRP} and FSDR \cite{Huang:2021:FSD} on most benchmarks (\eg, by $4.2-9.4$\% on Mapillary with ResNet-101).
These methods train individual models for each target domain; FSDR \cite{Huang:2021:FSD} resorts to the target domains for hyperparameter tuning, hence contravenes our proposed out-of-distribution evaluation protocol.
Recall that ASG \cite{Chen:2020:ASG} and CSG \cite{Chen:2021:CSG} \cite[as well as DRPC;][]{Yue:2019:DRP} require access to a distribution of real images for training, while IBN-Net \cite{Pan:2018:TAO} modifies the model architecture.
Our thoroughly straightforward approach requires neither, alters only the inference procedure, yet outperforms these methods in all benchmark scenarios substantially.
WildNet \cite{Lee:2022:CVPR} appears to be more accurate than self-adaptation on BDD with ResNet-101.
However, this is not a fair comparison, since it uses a more advanced architecture (DeepLabv3+ \textit{vs.} DeepLabv1).
As we will see in \cref{table:recent_arch}, self-adaptation, in fact, outperforms WildNet when using the same architecture by $2.79\%$ IoU.
Similarly, SAN-SAW \cite{Peng:2022:CVPR} reaches higher accuracy on \bdd, if trained on \synthia, presumably due to the ASPP module \cite{Chen:2018:DLS} that we do not use.
Self-adaptation considerably outperforms SAN-SAW in all other scenarios.
Overall, despite adhering to a stricter evaluation practice and a simpler model architecture, self-adaptation overwhelmingly exceeds the segmentation accuracy of previous work.

\myparagraph{Comparison to Tent~\citep{Wang:2020:FTT}.}
\begin{table}[t]
    \centering
    \caption{\textit{Mean IoU (\%,  $\uparrow$) comparison to Tent~\citep{Wang:2020:FTT}.} \emph{(a)} Using the same HRNet-W18 backbone \citep{Wang:2021:DHR} as in \citep{Wang:2020:FTT}, our self-adaptation outperforms Tent~\citep{Wang:2020:FTT} substantially, even when using SaN alone. \emph{(b)} Our self-adaptation loss defined in \cref{sec:sada} yields significantly higher segmentation accuracy compared to the entropy minimization used by Tent~\citep{Wang:2020:FTT}.}
    \subcaptionbox{HRNet-W18 backbone}{
    \label{table:hrnet-w18}
    \begin{tabularx}{0.33\linewidth}{@{}Xc@{}}
        \toprule
        Method & CS \\
        \midrule
        Tent~\citep{Wang:2020:FTT} & 36.4 \\
        \iabn \emph{(Ours)} & 40.0 \\
        Self-adaptation \emph{(Ours)} & \textbf{44.1} \\
        \bottomrule
    \end{tabularx}
}
\hfill
    \subcaptionbox{Comparison to entropy minimization (DeepLabv1)}{
    \begin{tabularx}{0.6\linewidth}{@{}Xcccc@{}}
        \toprule
        Method & CS & BDD & IDD & Mean \\
        \midrule
        Entropy min. & 40.20 & 35.40 & 33.97 & 36.52 \\
        \emph{Ours} ($\psi{=}0.0$)  & 44.00 & 38.55 & 39.09 & 40.55 \\
        \emph{Ours} ($\psi{=}0.7$) & \textbf{45.13} & \textbf{39.61} & \textbf{40.32} & \textbf{41.69} \\
        \bottomrule
    \end{tabularx}
}   
\label{tab:tent}
\end{table}
Like self-adaptation, Tent~\citep{Wang:2020:FTT} also updates model parameters at test time.
However, different from constructing the pseudo-labels based on well-calibrated predictions in our self-adaptation, Tent simply minimizes the entropy of a single-scale prediction.
Tent also limits the adaptation to updating only the BN parameters, whereas our self-adaptation generalizes this process to convolutional layers.
To demonstrate these advantages, we compare to Tent~\citep{Wang:2020:FTT} in \cref{tab:tent}.
We train HRNet-W18 \citep{Wang:2021:DHR} on GTA and compare the IoU on Cityscapes to the equivalent configuration of Tent.
Under a comparable computational budget of 10 model update iterations, we substantially outperform Tent, by a remarkable $7.5$\% IoU.
\iabn alone already outperforms Tent significantly with a single forward pass by 3.6\%.
Further, our self-adaptation loss is also considerably more effective than the entropy minimization employed by Tent. We improve by $4$\% IoU on average even without tuning threshold $\psi$, and by $5.2$\% when it is tuned.

\subsection{Additional analysis}
\label{sec:exp_aux}

\myparagraph{Runtime-accuracy trade-off.}
\begin{figure*}[t!]
    \includegraphics[width=\linewidth]{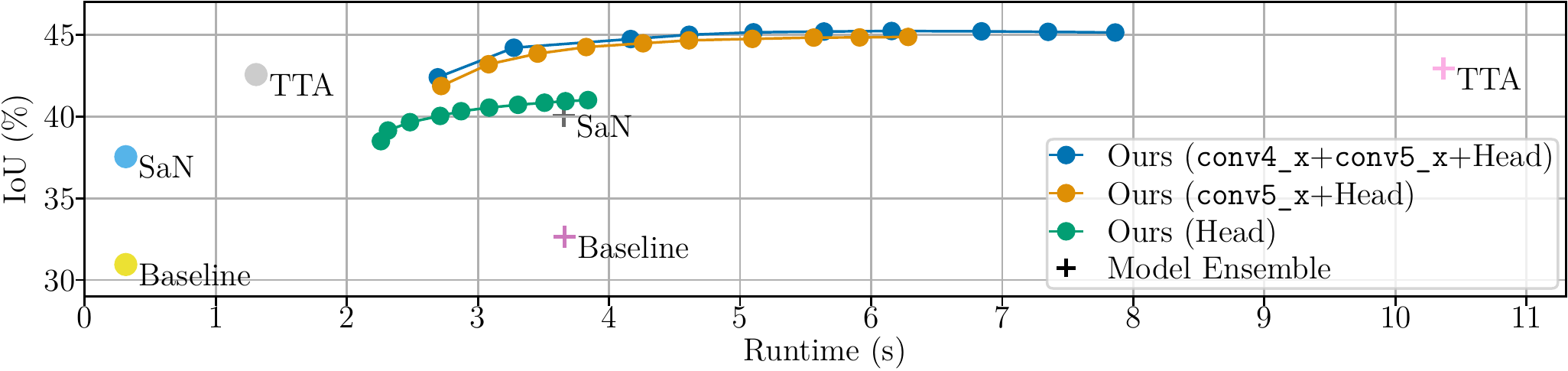}
    \caption{\textit{Runtime-accuracy comparison on \gta $\rightarrow$ Cityscapes generalization using one NVIDIA GeForce RTX 2080 GPU.} The curves trace self-adaptation iterations, \ie, the first point corresponds to $N_t=1$, while the last shows $N_t=10$. Self-adaptation balances accuracy and inference time by adjusting iteration numbers and layer choices, and is more cost-effective than 10 network ensembles.}
    \label{fig:acc_runtime}
\end{figure*}
We investigate the influence of the number of iterations required to adapt to a single sample during self-adaptation.
\cref{fig:acc_runtime} plots IoU scores for Cityscapes using the ResNet-50 backbone trained on GTA (\cref{sec:supp_ttt} provides a numerical comparison).
As a widely adopted baseline, we also train a model ensemble comprising 10 networks with a DeepLabv1 architecture (as in self-adaptation), initialized with a random seed \citep{Hansen:1990:NNE}.
Note that \tta, self-adaptation, and the ensemble use \iabn for a fair comparison, and we also test the ensemble with \tta.
Although self-adaptation increases the accuracy of the baselines at the expense of test-time latency, it is still more efficient and more accurate than the model ensembles.
Another advantage is that self-adaptation can trade off the accuracy \emph{vs.} runtime by using fewer update iterations, or updating fewer upper network layers.
While the top-accuracy variant of self-adaptation may not be suitable for real-time applications yet, it can still be valuable in other important domains, such as medical imaging, where high accuracy is desirable even at the cost of increased latency.
For real-time needs, \iabn alone boosts the baseline accuracy significantly without any computational overhead.

\begin{figure*}[t]
    \includegraphics[width=\linewidth]{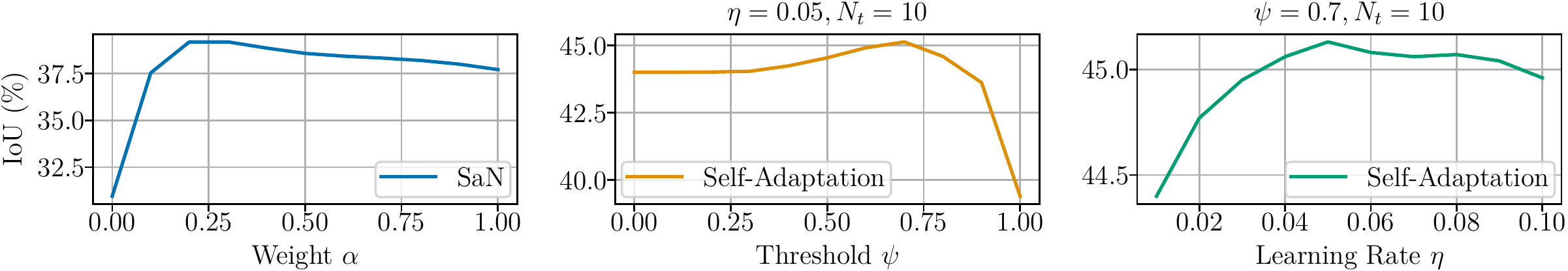}
    \caption{\textit{Hyperparameter sensitivity on \gta $\rightarrow$ Cityscapes generalization.} We investigate our hyperparameters $\alpha$, threshold $\psi$ and learning rate $\eta$ and report scores using Deeplabv1 with the ResNet-50 backbone trained on \gta. For self-adaptation, we fix one hyperparameter ($\psi$, $\eta$) while varying the other.}
    \label{fig:acc_hyperparams}
\end{figure*}

\myparagraph{Hyperparameter sensitivity}
\cref{fig:acc_hyperparams} plots segmentation accuracy \wrt our hyperparameters $\alpha$, $\psi$, and $\eta$. Complying with our protocol in \cref{sec:protocol}, we exclusively rely on our validation set to find the optimal values as follows. As detailed in \cref{sec:exp_iabn}, we first use SaN alone and determine $\alpha$ leading to the highest accuracy on the validation set, yielding $\alpha = 0.1$. Recall that the threshold $\psi$ plays a crucial role in determining the reliability of our pseudo-labels, while the learning rate $\eta$ governs the test-time adaptation process. We use grid search and fine-tune network parameters with self-adaptation on the validation set, which establishes the optimal values of $\psi = 0.7$ and $\eta = 0.05$. 
We observe from \cref{fig:acc_hyperparams} that the model accuracy on the validation set declines moderately as we deviate from the optimal hyperparameter values.
For example, if we were to increase $\eta$ twofold, the accuracy would drop by only $~0.15\%$ IoU.
\cref{sec:supp_ttt} provides numerical results to reproduce \cref{fig:acc_hyperparams}.
Overall, hyperparameters in self-adaptation exhibit a reasonable range of tolerance. 
Our comparision to the state of the art in \cref{sec:exp_iabn_ttt} also supports this conclusion, since these hyperparameter values may not be optimal for the target domains. Self-adaptation achieves superior segmentation accuracy nonetheless.

\myparagraph{Qualitative results.}
\begin{figure*}[t]%
    \def\svgwidth{1.0\linewidth}
    \input{figures/experiments/fresults_fig.tex}
    \caption{\textit{Qualitative semantic segmentation results for the generalization from \gta to Cityscapes, BDD, Mapillary, and IDD} for the ResNet-50 backbone. We show the input image (top row), ground truth and the predictions of the baseline model and of our proposed self-adaptation (bottom row).}
    \label{fig:qual_res}
    \vspace{-0.5em}
 \end{figure*}
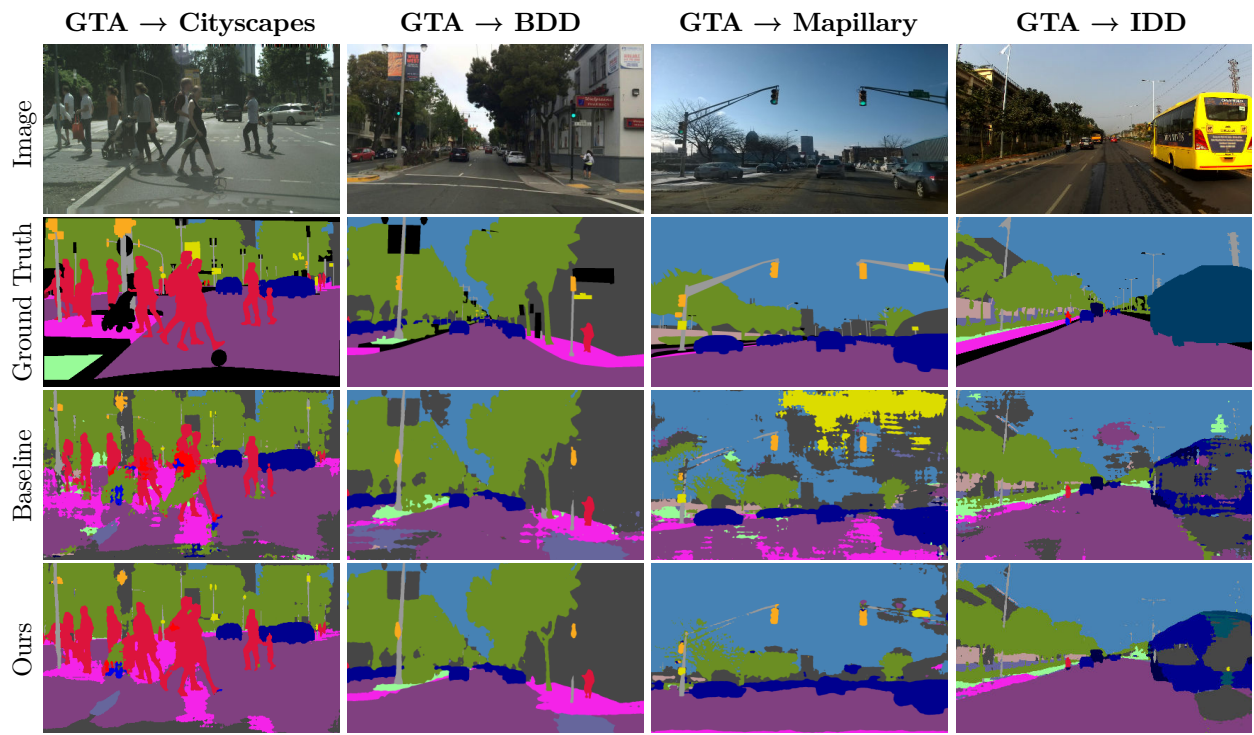
In \cref{fig:qual_res} we visualize qualitative segmentation results produced by self-adaptation for generalization from GTA to Cityscapes, BDD, Mapillary, and IDD.
We observe a clearly perceivable improvement over the baseline, especially in terms of consistency with the image boundaries.
\cref{sec:supp_qual} illustrates further examples and discusses failure cases.

%% file: figures/experiments/fresults_fig.tex
\begingroup%
  \makeatletter%
  \providecommand\color[2][]{%
    \errmessage{(Inkscape) Color is used for the text in Inkscape, but the package 'color.sty' is not loaded}%
    \renewcommand\color[2][]{}%
  }%
  \providecommand\transparent[1]{%
    \errmessage{(Inkscape) Transparency is used (non-zero) for the text in Inkscape, but the package 'transparent.sty' is not loaded}%
    \renewcommand\transparent[1]{}%
  }%
  \providecommand\rotatebox[2]{#2}%
  \newcommand*\fsize{\dimexpr\f@size pt\relax}%
  \newcommand*\lineheight[1]{\fontsize{\fsize}{#1\fsize}\selectfont}%
  \ifx\svgwidth\undefined%
    \setlength{\unitlength}{416.44803692bp}%
    \ifx\svgscale\undefined%
      \relax%
    \else%
      \setlength{\unitlength}{\unitlength * \real{\svgscale}}%
    \fi%
  \else%
    \setlength{\unitlength}{\svgwidth}%
  \fi%
  \global\let\svgwidth\undefined%
  \global\let\svgscale\undefined%
  \makeatother%
  \begin{picture}(1, 0.58789823)%
  \lsstyle
    \lineheight{1}%
    \setlength\tabcolsep{0pt}%
    \put(0,0){\includegraphics[width=\unitlength,page=1]{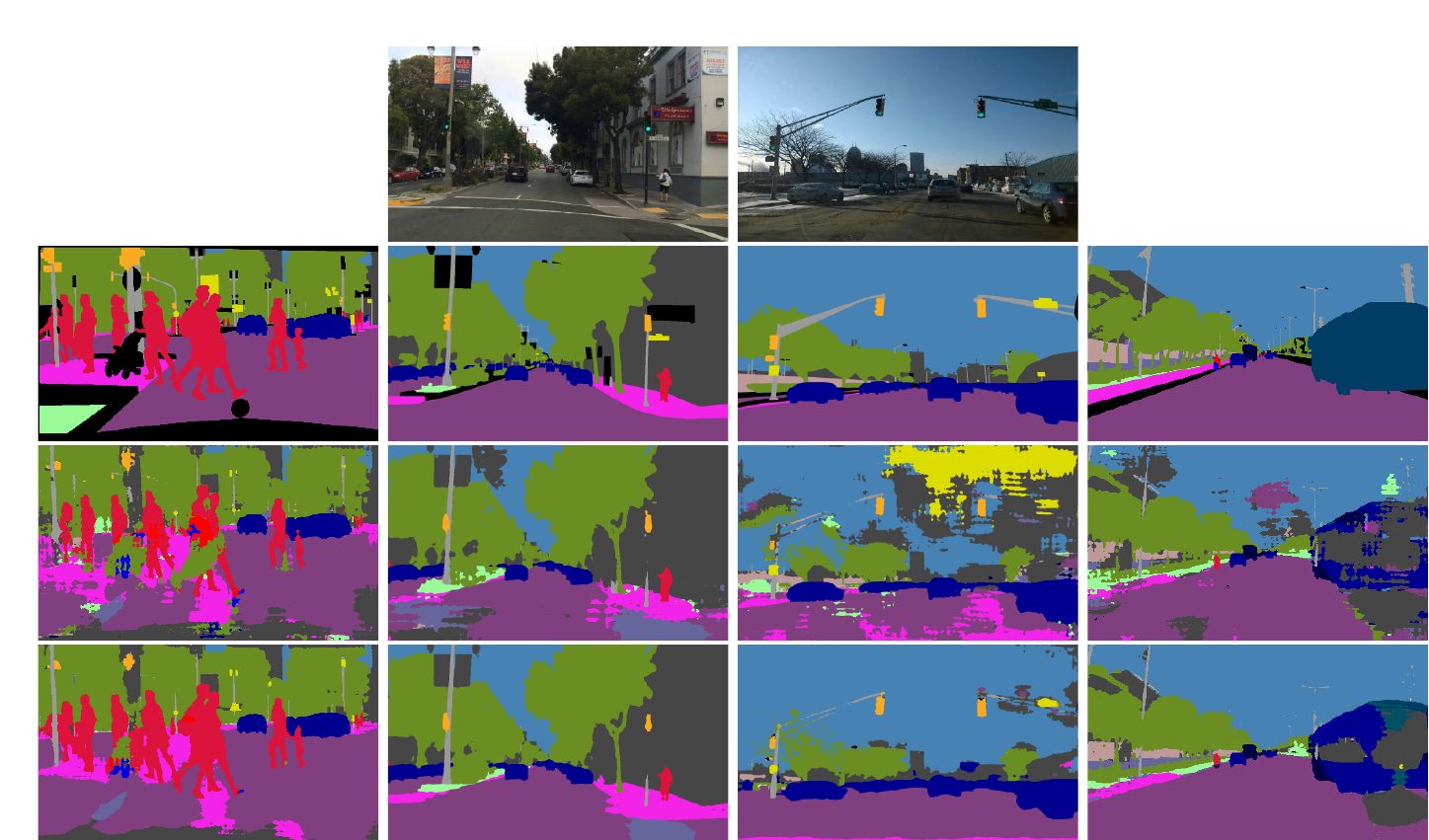}}%
    \put(0,0){\includegraphics[width=\unitlength,page=2]{figures/experiments/fresults_fig_im.pdf}}%
    \put(0.145,0.565){\makebox(0,0)[ct]{\lineheight{1.25}\smash{\begin{tabular}[t]{l}\textbf{GTA → Cityscapes} \end{tabular}}}}%
    \put(0.387,0.565){\makebox(0,0)[ct]{\lineheight{1.25}\smash{\begin{tabular}[t]{l}\textbf{GTA → BDD} \end{tabular}}}}%
    \put(0.635,0.565){\makebox(0,0)[ct]{\lineheight{1.25}\smash{\begin{tabular}[t]{l}\textbf{GTA → Mapillary} \end{tabular}}}}%
    \put(0.879,0.565){\makebox(0,0)[ct]{\lineheight{1.25}\smash{\begin{tabular}[t]{l}\textbf{GTA → IDD} \end{tabular}}}}%

    \put(0.017,0.455){\rotatebox{90}{\makebox(0,0)[lt]{\lineheight{1.25}\smash{\begin{tabular}[t]{l}Image \end{tabular}}}}}%
    \put(0.017,0.283){\rotatebox{90}{\makebox(0,0)[lt]{\lineheight{1.25}\smash{\begin{tabular}[t]{l}Ground Truth\end{tabular}}}}}%
    \put(0.017,0.17){\rotatebox{90}{\makebox(0,0)[lt]{\lineheight{1.25}\smash{\begin{tabular}[t]{l}Baseline \end{tabular}}}}}%
    \put(0.017,0.045){\rotatebox{90}{\makebox(0,0)[lt]{\lineheight{1.25}\smash{\begin{tabular}[t]{l}Ours\end{tabular}}}}}%
  \end{picture}%
\endgroup%

%% file: sections/conclusion.tex
 
The \emph{i.i.d.} assumption underlying the traditional learning principle, ERM, implies that a training process relying on it is unlikely to produce models robust to an arbitrary domain shift, unless we make further assumptions about the test distribution.
In the out-of-distribution scenario, the test domain is unknown, hence formulating such assumptions is difficult.
To bypass this issue, we presented and studied a self-adaptive \emph{inference} process.
We also highlighted a number of shortcomings in the experimental protocol used in previous work.
By following the best practice in machine learning research, we formulated four principles defining a rigorous evaluation process in domain generalization.
We implemented and followed these principles in our experiments. 

Our analysis demonstrates that a single sample from the test domain can already suffice to improve model predictions.
The accuracy improvement shown by our experiments is surprisingly substantial, despite using a fairly straightforward approach without changes to the training process or the model architecture, unlike in previous works \citep{Chen:2020:ASG,Yue:2019:DRP}.
We hope that these encouraging results will incentivize our research community to study self-adaptive techniques in other application domains, such as panoptic segmentation, or monocular depth prediction.

The particular instantiation of self-adaptation that we presented in this work is not yet real-time.
We extensively analyzed the existing trade-off with the segmentation accuracy in \cref{sec:exp_aux} and \cref{sec:runtime}, and found self-adaptation to be nonetheless more cost-effective than model ensembles --- a widely adopted framework in machine learning.
However, the scope of this work was to rigorously demonstrate concrete accuracy benefits, which will make improving model efficiency a worthwhile goal to pursue in future work.

Indeed, decreasing the latency of self-adaptive inference is an intriguing avenue for research.
Using adaptive step sizes, higher-order optimization \citep{Botev:2017:PGN,Kingma:2015:AAM}, or implicit gradients \citep{Rajeswaran:2019:MLW}, we may lower the number of required update steps.
An orthogonal line of research may consider low-precision computations \citep{Wang:2019:E2T}, low-rank decomposition \citep{Jaderberg:2014:SUC}, or alternating update strategies \citep{Tonioni:2019:RTS} to offer reduced computational footprint for each update iteration.
We are excited to explore these directions and hope the reader may find this work equally inspiring.

%% file: supp/sections/supp_baseline.tex
To legitimately study the out-of-distribution model accuracy, it is essential to establish its upper bound attainable by the standard training procedures \citep{Montavon:2012:NNT}.
We followed the best practice in the literature and found a number of training details to be crucial for obtaining a highly competitive baseline, \ie, a model that does not use our self-adaptive inference.
Among them is using heavy data augmentation.
Recall from \cref{sec:exp} that we used random horizontal flipping, multi-scale cropping with a scale range of $[0.08, 1.0]$, as well as photometric image perturbations:\footnote{We use Pillow library (\href{https://pillow.readthedocs.io}{https://pillow.readthedocs.io}) to implement photometric augmentation.} color jitter, random blur, and grayscaling.
Color jitter, applied with probability 0.5, perturbs image brightness, contrast, and saturation using a factor sampled uniformly from the range $[0.7, 1.3]$.
We use a different range of $[0.9, 1.1]$ for the hue factor.
We randomly blur the image using a Gaussian kernel with the standard deviation sampled from $[0.1,2.0]$.
Additionally, we convert the image to grayscale with a probability of $0.1$.
Furthermore, we also found that the polynomial decay schedule we used for the learning rate, as well training for at least 50 epochs (for both \gta and \synthia) are essential to achieve a high baseline accuracy.
Note that we only used \wilddash as the development set to tune these training details.
We also experimented with higher input resolution and a larger batch size, but did not observe a significant improvement, yet a drastic increase in the computational overhead.

\myparagraph{On importance of the baseline.}
We note that the reported accuracy from previous work in \cref{table:sota_cmp} is not entirely consistent \wrt\ the choice of the model architecture.
In particular, DRPC \citep{Yue:2019:DRP} 
uses an FCN, yet outperforms other domain generalization approaches with DeepLabv1 \citep{Pan:2018:TAO} 
and DeepLabv2 \citep{Chen:2020:ASG,Chen:2021:CSG} 
considerably, as shown in \cref{table:sota_cmp}.
This is a regrettable consequence of inconsistent training schedules used in previous works that proved difficult to reproduce.
For example, at the time of submission, the implementation by Yue~\etal \citep{Yue:2019:DRP}, 
which reports excellent segmentation accuracy for the baseline (\cf \cref{table:sota_cmp}), has not yet been made available;\footnote{\href{https://github.com/xyyue/DRPC/issues}{https://github.com/xyyue/DRPC/issues}} parts of the code implementing the semantic segmentation architecture introduced in \citep{Pan:2018:TAO} are also not publicly available.\footnote{\href{https://github.com/XingangPan/IBN-Net}{https://github.com/XingangPan/IBN-Net}}
These circumstances make reporting the accuracy of the implementation-specific baselines indispensable, which has thus become the standard practice in more recent previous \citep{Chen:2020:ASG,Chen:2021:CSG} and related works \citep{Gulrajani:2020:ISL}.

%% file: supp/sections/supp_SaN.tex
\begin{figure*}[t]%
    \includegraphics[width=1.0\linewidth]{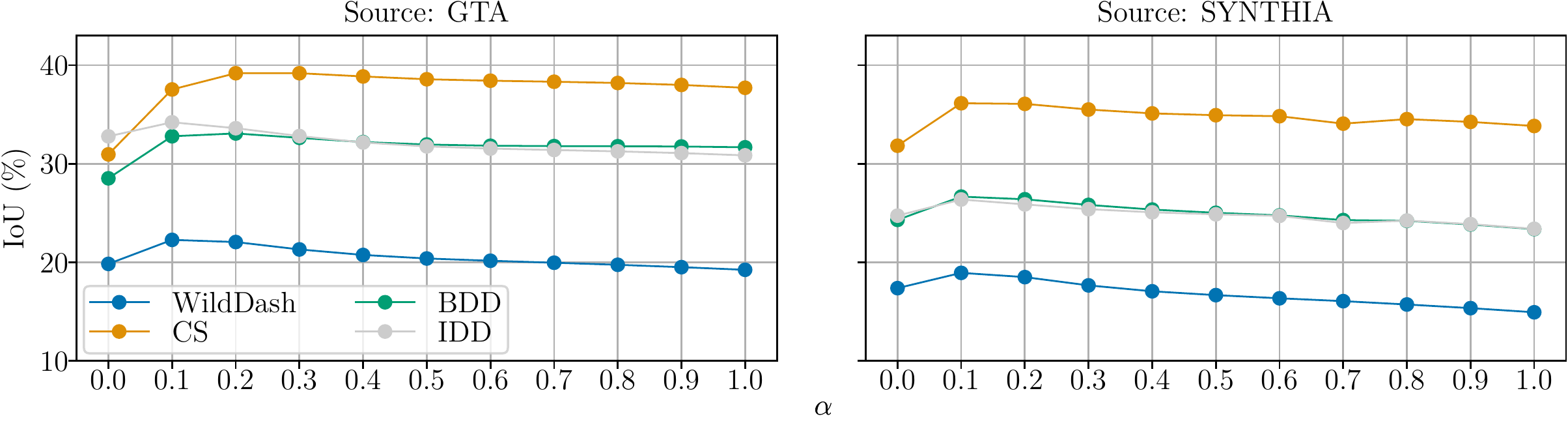}
    \caption{\textit{Mean IoU (\%, $\uparrow$) using \iabn based on the optimal alpha on the development set (\wilddash).} We report scores for the target domains (\cityscapes, \bdd, \idd) for the ResNet-50 backbone after training on \gta \emph{(left)} and \synthia \emph{(right)}.}
    \label{fig:alpha_resnet50}
\end{figure*}

\myparagraph{Selecting \boldsymbol{$\alpha$}.}
\cref{fig:alpha_resnet50} shows a detailed plot of the influence of $\alpha$ on the segmentation accuracy, both on the development set of \wilddash and on the target domains.
We observe that the maximum accuracy on the development set is attained with $\alpha=0.1$.
Clearly, there is no guarantee that value $0.1$ is the optimal one for the target domains.
However, choosing $\alpha$ based on the development set is in line with the established practice in machine learning: Tuning model hyperparameters is not allowed on the test sets (\ie, Cityscapes, \bdd, \idd), but is only possible on the validation set (\wilddash).
In general, the hyperparameters found to be optimal on the validation set are not guaranteed to remain so on the test set, especially in the out-of-distribution scenario studied here.
Nevertheless, self-adaptation is quite robust even to the inevitably suboptimal choice of the hyperparameters in the out-of-distribution setting.
Despite $\alpha$ having been picked based on the validation dataset, our empirical results show consistent improvements over the baselines across all scenarios (\cf \cref{table:TTTvsTTA}).

\myparagraph{\synthia as the training set.}
Due to space constraints, we limited our study of \iabn in the main paper to the scenario of using \gta as the source data (\cf \cref{table:calibration}).
Here, we extend this study by training our models on \synthia instead.
\cref{table:iabn_synthia} reports the segmentation accuracy (in terms of IoU) and the expected calibration error (ECE) for this case.
Notably, \iabn can still provide benefits for the expected segmentation accuracy if \pbn (\ie, using target instance normalization statistics) fails to improve over the \tbn baseline, as is the case with ResNet-50 in \cref{table:iabn_tbn_pbn_synthia};
the results remain on par with the \tbn baseline even when \pbn is significantly worse than \tbn.
In regard to calibration quality, the results are consistent with our model trained on \gta (\cf \cref{table:iabn_calibration}(b)): Not only does \iabn improve the prediction calibration of the baseline, it again exhibits a complementary effect with Monte Carlo dropout \citep{Gal:2016:DBA}.
Overall, the combined results from \cref{table:iabn_calibration} and \cref{table:iabn_synthia} demonstrate that \iabn improves both the model accuracy and the calibration quality of the predictions in the out-of-distribution setting irrespective of the backbone network and specifics of the source data.

\begin{table}[!t]
    \caption{\textit{Mean IoU (\%, $\uparrow$) comparison of \iabn to alternatives: SN~\citep{Luo:2019:DLN} and BIN~\citep{Nam:2018:BIN}.}}
    \label{table:exp_norm}
    \begin{tabularx}{\linewidth}{@{}Xcccc@{}}
        \toprule
        Method & CS & BDD & IDD & Mean \\
        \midrule
        SN & 31.75 & \textbf{33.60} & 31.60 & 32.32 \\
        BIN & 34.57 & 32.68 & 30.22 &  32.49 \\
        \iabn \emph{(ours)} & \textbf{37.54} & 32.79 & \textbf{34.21} & \textbf{34.85} \\
        \bottomrule
    \end{tabularx}
\end{table}

\myparagraph{Comparison to other related work.}
We additionally compare \iabn to alternative normalization strategies proposed in the literature: Batch-Instance Normalization \citep[BIN;][]{Nam:2018:BIN} and Switchable Normalization \citep[SN;][]{Luo:2019:DLN}.
Although both of these techniques share some similarities with our \iabn, these approaches were developed for different purposes.
SN was only shown to improve in-domain accuracy, while BIN tackles domain adaptation for image classification, not domain generalization studied in this work.
Furthermore, both methods modify the model architecture before training, while \iabn works with any pretrained semantic segmentation model.
Nevertheless, we implemented both SN and BIN in our segmentation model based on the ResNet-50 backbone.
We trained these approaches on GTA in an identical setup as \iabn.
From results in \cref{table:exp_norm}, we observe that \iabn outperforms both BIN and SN by a significant margin in terms of mean IoU of the target domains.

\begin{table}[!t]
    \caption{\emph{(a)} \textit{Segmentation accuracy using \iabn.} We report the mean IoU(\%, $\uparrow$) on three target domains (Cityscapes, BDD, IDD) across both backbones. In contrast to \cref{table:iabn_calibration}, we trained the networks on \synthia in both cases. As before, \tbn denotes train BN \citep{Ioffe:2015:BNA}, 
    while \pbn refers to prediction-time BN \citep{Nado:2020:EPT}.
    \emph{(b)} \textit{ECE for \iabn and MC-Dropout \citep{Gal:2016:DBA}}. We report ECE scores (\%, $\downarrow$) for three target domains (Cityscapes, BDD, IDD) across both backbones.
    }
    \label{table:iabn_synthia}
    \centering
    \vspace{-0.5em}
    \subcaptionbox{}{%
     \setlength\tabcolsep{2pt}
        \begin{tabularx}{0.47\linewidth}{@{}Xcccc@{}}
        \toprule
        \multirow{2}{*}{Method} & \multicolumn{4}{@{}c}{\textit{IoU (\%, $\uparrow$)}}\\
        \cmidrule(l){2-5}
        & CS & BDD  & IDD & Mean \\
        \midrule
        ResNet-50 & & & & \\
        \ \ w/ \tbn & 31.83 & 24.30 & 24.73 & 26.95 \\
        \ \ w/ \pbn & 33.83 & 23.36 & 23.39 & 26.86 \\
        \ \ w/ \iabn \emph{(Ours)} & \textbf{36.14} & \textbf{26.66} & \textbf{26.37} & \textbf{29.72} \\
        \midrule
        ResNet-101 & & & & \\
        \ \ w/ \tbn & 37.25 & \textbf{29.32} & 27.19 & 31.25 \\
        \ \ w/ \pbn & 34.58 & 24.24 & 22.32 & 27.05\\
        \ \ w/ \iabn \emph{(Ours)} & \textbf{38.01} & 28.66 & \textbf{27.28} & \textbf{31.32}\\
        \bottomrule
      \end{tabularx}
        \label{table:iabn_tbn_pbn_synthia}
    }\hfill
    \subcaptionbox{}{%
    \setlength\tabcolsep{2pt}
    \begin{tabularx}{0.47\linewidth}{@{}X@{}cccc@{}}
        \toprule
        \multirow{2}{*}{Method} & \multicolumn{4}{@{}c}{\textit{ECE (\%, $\downarrow$)}}\\
        \cmidrule{2-5}
         & CS & BDD  & IDD & Mean \\
        \midrule
        ResNet-50 & 37.50 & 43.19 & 40.11 & 40.26 \\
        \ \ w/ \iabn & 30.96 & 33.27 & 36.31 & 33.51\\
        \ \ w/ MC-Dropout  & 34.82 & 37.30 & 36.63 & 36.25 \\
        \ \ w/ both (\emph{Ours}) & \textbf{30.66} & \textbf{33.06} & \textbf{35.60} & \textbf{33.11} \\
        \midrule
        ResNet-101 & 31.39 & 33.77 & 36.56 & 33.91\\
        \ \ w/ \iabn & 30.33 & 31.83 & 36.26 & 32.81\\
        \ \ w/ MC-Dropout & 32.73 & 32.76 & 34.07 & 33.19 \\
        \ \ w/ both (\emph{Ours}) & \textbf{27.71} & \textbf{30.48} & \textbf{32.67} & \textbf{30.29} \\
        \bottomrule
    \end{tabularx}
    \label{table:calibration_synthia}
    }
\end{table}

\begin{table}[!t]
    \caption{\textit{Mean IoU (\%, $\uparrow$) for \iabn and MC-Dropout \citep{Gal:2016:DBA}}. We report scores for three target domains (Cityscapes, BDD, IDD) on the ResNet-50 \citep{He:2016:DRL} backbone trained on \gta. We provided the ECE results in \cref{table:iabn_calibration}(b) to motivate our self-adaptive learning approach, which relies on improved model calibration. \iabn also consistently improves the IoU of the MC-Dropout approach.}
    \label{table:iabn_tbn_pbn_synthia_cal}
    \centering
    \begin{tabularx}{\linewidth}{@{}Xcccc@{}}
        \toprule
        \multirow{2}{*}{Method} & \multicolumn{4}{@{}c}{\textit{IoU (\%, $\uparrow$)}}\\
        \cmidrule(l){2-5}
        & CS & BDD  & IDD & Mean \\
        \midrule
        ResNet-50 & 30.95 & 28.52 & 32.78 & 30.75\\
        \ \ w/ \iabn \emph{(Ours)} & 37.54 & 32.79 & 34.21 & 34.85 \\
        \ \ w/ MC-Dropout & 30.45 & 31.96 & 32.50 & 31.63 \\
        \ \ w/ both \emph{(Ours)} & \textbf{38.84} & \textbf{35.13} & \textbf{35.55} & \textbf{36.50} \\
        \bottomrule
      \end{tabularx}

\end{table}

%% file: supp/sections/supp_SA.tex
\myparagraph{Selecting parameters for self-adaptation.}
In self-adaptation, we only update the model parameters in the layers \texttt{conv4\_x}, \texttt{conv5\_x}, and the classification head for the ResNet-50 backbone. Due to the higher computational cost of the ResNet-101 backbone, we only adapt the layers \texttt{conv5\_x} and the classification head in this case.
Note that we analyze the runtime costs for alternative configurations in \cref{fig:acc_runtime} in the main text, as well as \cref{fig:acc_runtime_other} below.

\myparagraph{Threshold and learning rate.}
We investigate the influence of hyperparameters for self-adaptation: threshold $\psi$ and learning rate $\eta$. \cref{table:hyperparameters} reports the IoU after self-adaptation evaluated on the target domain Cityscapes after training with a ResNet-50 backbone on GTA. As already observed for $\alpha$ (\cf \cref{fig:alpha_resnet50}), self-adaptation is robust to the choice of hyperparameters.

\begin{table}[!t]
    \caption{\textit{Mean IoU (\%, $\uparrow$) for hyperparameter variation of threshold $\psi$ and learning rate $\eta$.} We report scores on the ResNet-50 backbone trained on GTA evaluated on Cityscapes for self-adaptation and SaN for reference.}
    \label{table:hyperparameters}
    \setlength\tabcolsep{3pt}
    \begin{tabularx}{\linewidth}{@{}Xcccccccccccc@{}}
        \toprule
        &  &\multicolumn{11}{@{}c}{Self-Adaptation \textit{(Ours): variation of $\psi$ ($\eta = 0.05$, $N_t = 10$)}} \\
        \cmidrule(l){3-13}
        & w/ SaN \textit{(Ours)} & 0.0 & 0.1 & 0.2 & 0.3 & 0.4 & 0.5 & 0.6 & 0.7 & 0.8 & 0.9 &1.0  \\
        \midrule
        ResNet-50 &37.54 &44.00& 44.00& 44.01& 44.04& 44.24& 44.54& 44.91& \textbf{45.13} & 44.59& 43.62& 39.40 \\
        \midrule
        &  & &\multicolumn{10}{@{}c}{Self-Adaptation \textit{(Ours): variation of $\eta$ ($\psi = 0.7$, $N_t = 10$)}} \\
        \cmidrule(l){4-13}
        & w/ SaN \textit{(Ours)} & & 0.01     &     0.02    &0.03     &0.04     &0.05     &0.06    &0.07     &0.08     &0.09     &0.1  \\
        \midrule
        ResNet-50 &37.54 & &44.40 &44.77 &44.95 &45.06 & \textbf{45.13} &45.08 &45.06 &45.07 &45.04 &44.96 \\ 
        \bottomrule
    \end{tabularx}
\end{table}

\begin{table}[!t]
    \caption{\textit{The role of the augmentation type in self-adaptation.} We report mean IoU (\%, $\uparrow$) and runtime (ms, $\downarrow$) for \tta \citep{Simonyan:2015:VDC} and our self-adaptation for the \gta source domain and the Cityscapes target domain for the ResNet-50 backbone.}
    \smallskip
    \centering
    \begin{tabularx}{\linewidth}{@{}Xcc@{\hspace{1em}}cc@{}}
        \toprule
        \multirow{2}{*}{Method}  & \multicolumn{2}{c@{\hspace{1em}}}{\textit{\tta}} & \multicolumn{2}{c@{}}{\textit{Ours}} \\
        \cmidrule(lr{1em}){2-3} \cmidrule{4-5}
        & IoU & Runtime & IoU & Runtime \\
        \midrule Baseline & 30.95 & \textbf{314} & 31.47 & \textbf{7135} \\
        \iabn & 37.54 & \textbf{314} & 39.04 & \textbf{7135} \\
        \midrule
        Multiple scales & 42.27 & 749 & 44.92 & 7272 \\
        Horizontal flipping & 38.01 & 986 & 39.33 & 7593 \\
        Grayscaling & 37.96 & 908 & 39.65 & 7193 \\
        \midrule
        Multiple scales + horizontal flipping & 42.48 & 1316 & 44.94 & 7890 \\
        Multiple scales + grayscaling & 42.28 & 1202 & 45.06 & 7486 \\
        Multiple scales + horizontal flipping + grayscaling & \textbf{42.56} & 1308 & \textbf{45.13} & 7862 \\
        \bottomrule
    \end{tabularx}
    \label{fig:abl_augm}
\end{table}

\myparagraph{The choice of augmentation strategies.}
We verify the influence of the augmentation type used by self-adaptation.
Recall from \cref{sec:exp_ttt} that we use multiple scales with horizontal flipping and grayscaling to augment one image sample.
We compare a flipping-only, scaling-only, and grayscaling-only version of our self-adaptation to the combination of flipping, grayscaling, and the spatial scaling, which we used in the main text.
We used a ResNet-50 backbone trained on GTA and report the accuracy on Cityscapes in \cref{fig:abl_augm}. We observe a significant boost in accuracy in comparison to a strong baseline that uses our \iabn with no augmentations.
Furthermore, we show that using multiple scales is more important than flipping for self-adaptation.
Note that the augmentations used do not impact the runtime in a significant way, since the batch sizes between these setups vary insignificantly; it is the backpropagation that dominates the main computational footprint.
Varying the number of iterations, as studied in \cref{sec:exp_iabn}, provides a more flexible mechanism for accuracy-runtime trade-off.

\myparagraph{Inference time.}
\label{sec:runtime}
\cref{table:TTTvsTTA_runtime} compares the inference time of our self-adaptation \wrt\ test-time augmentation (TTA) and single-scale inference across a range of input resolutions available in the target domains.
We obtain these results by running inference on a single NVIDIA GeForce RTX 2080 GPU.
To improve the runtime estimate, for each dataset we average the inference time over the complete image set.
This is to account for small deviations in the input resolution (\eg, IDD mostly contains images of resolution $1920 \times 1080$, but also has images with resolution $1280 \times 720$).
Since our self-adaptation uses 10 update iterations, the increase in the inference time \wrt\ TTA is expected.
Although such cost may be detrimental for real-time applications, the significant accuracy benefits of self-adaptation (\cf \cref{table:sota_cmp}) may find appeal in applications where the importance of the prediction quality outweighs the incurred overhead in the frame rate (\eg, medical image analysis).
Furthermore, self-adaption can amortize the runtime costs by using fewer update iterations, while still providing a clear accuracy improvement.
For example, using $N_t=3$ iterations decreases the runtime of self-adaptation almost twofold while preserving around $95$\% of the model accuracy attainable with more iterations.

\begin{table}[!t]
    \caption{\textit{Runtime with \tta \citep{Simonyan:2015:VDC} or self-adaptation.} We report the runtime (ms, $\downarrow$) for both ResNet-50 and ResNet-101 on three dominant resolutions of $2048 \times 1024$, $1280 \times 720$, and $1920 \times 1080$, corresponding to the target domains Cityscapes, BDD, and IDD, respectively.}
    \label{table:TTTvsTTA_runtime}
    \smallskip
    \centering
    \begin{tabularx}{\linewidth}{@{}Xcccc@{}}
        \toprule
        \multirow{2}{*}{Method}  & \multicolumn{4}{c@{}}{\textit{Input resolution}} \\
        \cmidrule(lr){2-5}
        & CS & BDD & IDD & Mean \\
        \midrule
        ResNet-50 (w/ \iabn) & \textbf{314} & \textbf{136} & \textbf{214} & \textbf{221} \\
        \ \ \tta (w/ \iabn) & 1308 & 713 & 766 & 929 \\
        \ \ Self-adaptation (\emph{ours}) & 7862 & 3742 & 5307 & 5637 \\
        \midrule
        ResNet-101 (w/ \iabn) & \textbf{458} & \textbf{239} & \textbf{252} & \textbf{316} \\
        \ \ \tta (w/ \iabn) & 1519 & 766 & 860 & 1048 \\
        \ \ Self-adaptation (\emph{ours}) & 9060 & 4241 & 6142 & 6481 \\
        \bottomrule
    \end{tabularx}
\end{table}

We also investigated the influence of using the automatic mixed precision module in PyTorch and its influence on the inference runtime.
While maintaining an identical IoU on Cityscapes using the ResNet-50 backbone, mixed precision with $N_t=10$ reduced the runtime by almost $30$\%: the inference takes only $5746$ ms compared to $7862$ ms using single precision.

\begin{table}[!t]
    \caption{\textit{Mean IoU (\%, $\uparrow$) using our self-adaptation integrated with DeepLabv3+ \citep{Chen:2018:ECA} based on a ResNet-50 and ResNet-101 backbone as well as with HRNet-W18, HRNet-W48 \citep{Wang:2021:DHR}, and UPerNet \citep{Xiao:2018:Unified} with a Swin-T backbone \citep{Liu:2021:Swin}}. We observe substantial improvements of the segmentation accuracy on all three target domains (Cityscapes, BDD, and IDD) after training on GTA.}
    \label{table:recent_arch}
    \smallskip
    \centering
    \begin{tabularx}{\linewidth}{@{}Xcccc@{}}
        \toprule
        \multirow{2}{*}{Method}  & \multicolumn{4}{c@{}}{\textit{Target domains}} \\
        \cmidrule(lr){2-5}
        & CS & BDD & IDD & Mean \\
        \midrule
        DeepLabv3+ ResNet-50 (Baseline) & 37.51 & 35.45 & 37.50 & 36.82 \\
        \ \ Self-adaptation \emph{(ours)} & \textbf{46.56} & \textbf{43.17} & \textbf{44.07} & \textbf{44.60} \\
        \midrule
        DeepLabv3+ ResNet-101 (Baseline) & 38.19 & 37.05 & 38.22 & 37.82 \\
        \ \ Self-adaptation \emph{(ours)} & \textbf{48.14} & \textbf{44.52} & \textbf{45.72} & \textbf{46.13} \\
        \midrule
        HRNet-W18 (Baseline) & 33.08 & 29.40 & 32.97 & 31.82 \\
        \ \ Self-adaptation \emph{(ours)} & \textbf{44.05} & \textbf{38.29} & \textbf{43.78} & \textbf{42.04} \\
        \midrule
        HRNet-W48 (Baseline) & 34.66 & 30.85 & 34.64 & 33.38 \\
        \ \ Self-adaptation \emph{(ours)} & \textbf{48.82} & \textbf{42.79} & \textbf{43.74} & \textbf{45.12} \\
        \midrule
        UPerNet Swin-T (Baseline) & 39.67 & 36.04 & 39.74 & 38.48 \\
        \ \ Self-adaptation \emph{(ours)} & \textbf{45.04} & \textbf{39.77} & \textbf{44.33} & \textbf{43.05} \\
        \bottomrule
    \end{tabularx}
\end{table}

We additionally provide runtime-accuracy plots for \gta $\rightarrow$ BDD and \gta $\rightarrow$ IDD generalization in \cref{fig:acc_runtime_other}.
The data supports our conclusions drawn on \gta $\rightarrow$ Cityscapes generalization (\cf \cref{sec:exp_iabn_ttt}) that \textit{(i)} self-adaptation provides clear advantages in segmentation accuracy over baselines at a reasonable increase of the inference time; \textit{(ii)} it is both more accurate and more efficient than model ensembles; and \textit{(iii)} it exhibits a flexible runtime-accuracy trade-off by means of varying the number of update iterations and the number of the layers to adjust.

\begin{figure*}[t]
    \begin{subfigure}{\linewidth}
        \includegraphics[width=\linewidth]{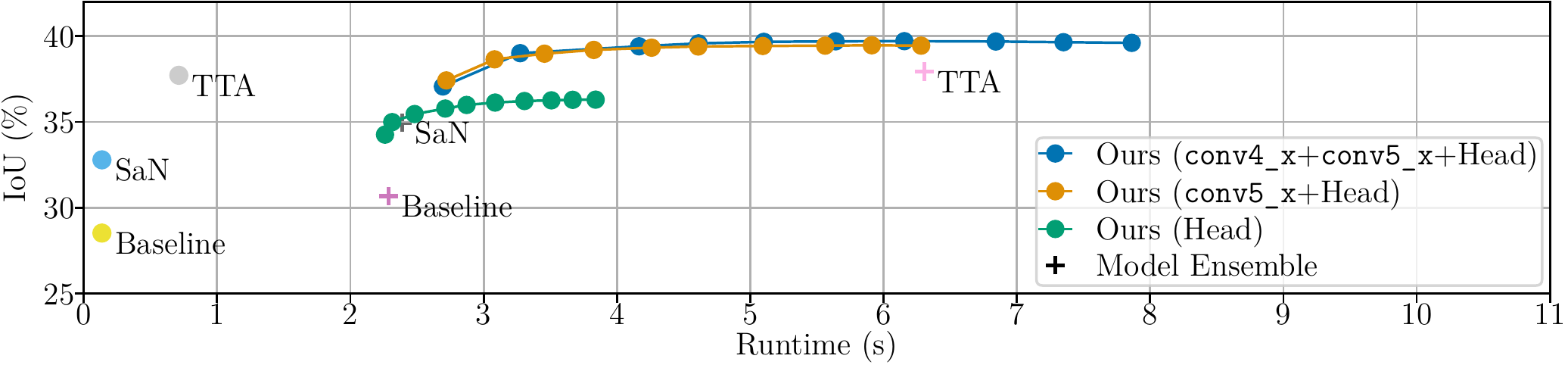}
    \end{subfigure}

    \vspace{10pt}
    \begin{subfigure}{\linewidth}
        \includegraphics[width=\linewidth]{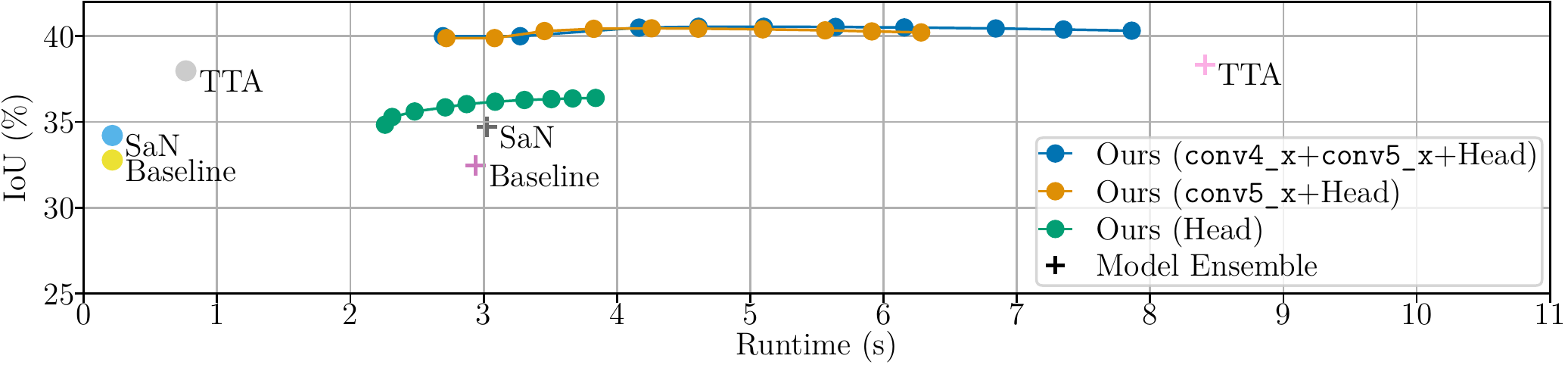}

    \end{subfigure}
    \caption{\textit{Runtime-accuracy comparison on \gta $\rightarrow$ BDD (top) and \gta $\rightarrow$ IDD (bottom) generalization using ResNet-50.} The curves on this plot trace self-adaptation iterations, \ie the first point corresponds to $N_t=1$, while the last shows $N_t=10$. While self-adaptation increases the inference time of the baseline and \tta for the sake of improved accuracy, it is still more efficient and accurate than model ensembles of 10 networks. The choice of the layers for self-adaptation updates \citep[the naming follows][]{He:2016:DRL} further provides a favorable runtime-accuracy trade-off. The runtime is computed on a single NVIDIA GeForce RTX 2080 GPU.}
    \label{fig:acc_runtime_other}
\end{figure*}

%% file: supp/sections/supp_arch.tex
Our approach generalizes to more recent architectures.
We trained five state-of-the-art segmentation models on GTA: DeepLabv3+ \citep{Chen:2018:ECA} with both a ResNet-50 and ResNet-101 backbone, HRNet-W18 and HRNet-W48 \citep{Wang:2021:DHR}, as well as UPerNet \citep{Xiao:2018:Unified} with a Swin-T Transformer backbone \citep{Liu:2021:Swin}. 
\cref{table:recent_arch} reports consistent and substantial improvement of the mean IoU over the baseline, across all these architectures and the target domains.

%% file: supp/sections/supp_diversedomains.tex
The ACDC dataset~\citep{Sakaridis:2021:ACDC} offers densely labeled driving scenes under adverse weather conditions such as fog, rain, and snow. As far as we are aware, previous work on domain generalization for semantic segmentation did not consider this benchmark. Therefore, we are only able to compare our results to our own baseline.\\
We evaluate our Deeplabv1 model with a ResNet-50 backbone trained on GTA with and without our self-adaptation approach. \cref{table:diversedomains} presents the results. Our method demonstrates a stark improvement across all weather conditions outperforming our baseline model by $13.57\%$ on average. This experiment underscores strong multi-target domain generalization of self-adaptation, since we evaluate the same model as we did for other target datasets without any change (\cf \cref{table:sota_cmp}). 
\begin{table}[!t]
    \caption{\textit{Mean IoU (\%, $\uparrow$) using our self-adaptation integrated with DeepLabv1 based on a ResNet-50 backbone.} We observe substantial improvements of the segmentation accuracy under adverse weather conditions (fog, rain, and snow) on the ACDC \citep{Sakaridis:2021:ACDC} validation set after training on the synthetic GTA.}
    \label{table:diversedomains}
    \smallskip
    \centering
    \begin{tabularx}{\linewidth}{@{}Xcccc@{}}
        \toprule
        \multirow{2}{*}{Method}  & \multicolumn{4}{c@{}}{\textit{Weather conditions}} \\
        \cmidrule(lr){2-5}
        & Fog & Rain & Snow & Mean \\
        \midrule
        ResNet-50 (Baseline) & 26.12 & 27.32 & 24.64 & 26.03 \\
        \ \ Self-adaptation \emph{(ours)} & \textbf{41.61} & \textbf{37.54} & \textbf{39.66} & \textbf{39.60} \\
        \bottomrule
    \end{tabularx}
\end{table}

%% file: supp/sections/supp_qualitative.tex
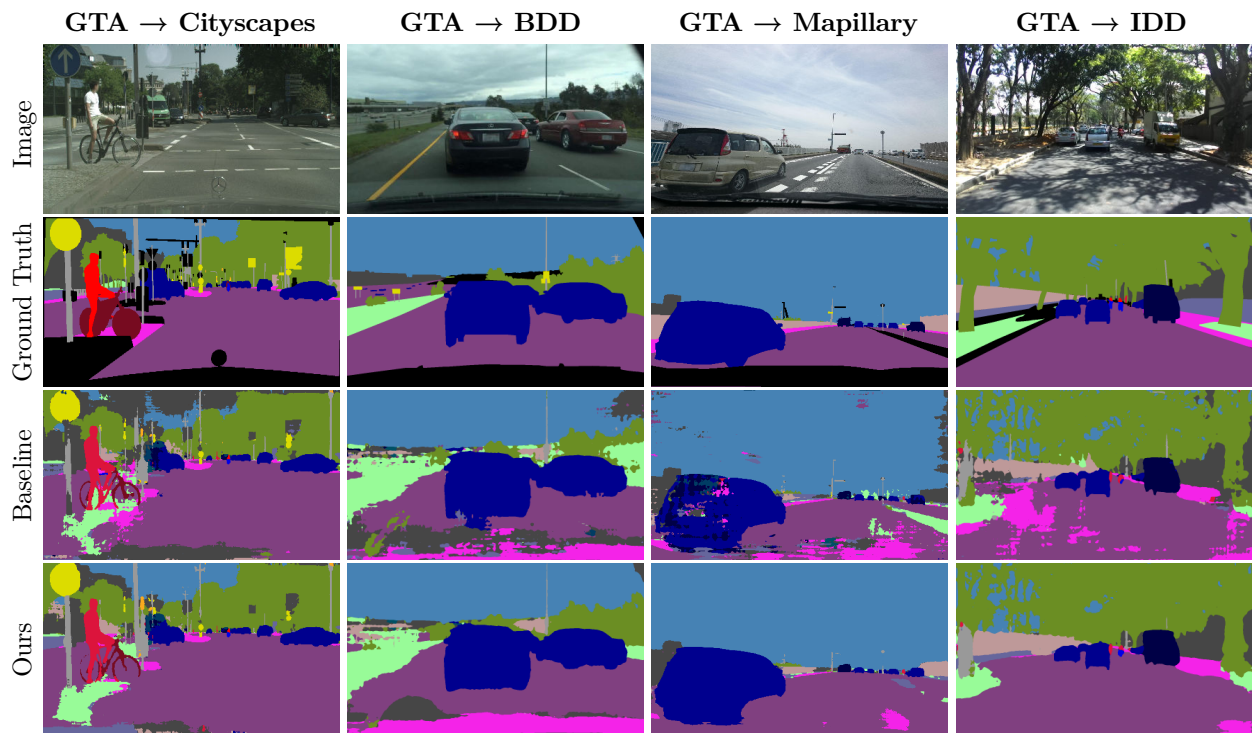
\begin{figure*}[!t]%
    \def\svgwidth{1.0\linewidth}
    \input{supp/figures/fresults_fig_gta101.tex}
    \caption{\textit{Qualitative semantic segmentation results for generalization from \gta to Cityscapes, BDD, Mapillary, and IDD} for the ResNet-101 backbone.}
    \label{fig:qual_res_gta101}
 \end{figure*}
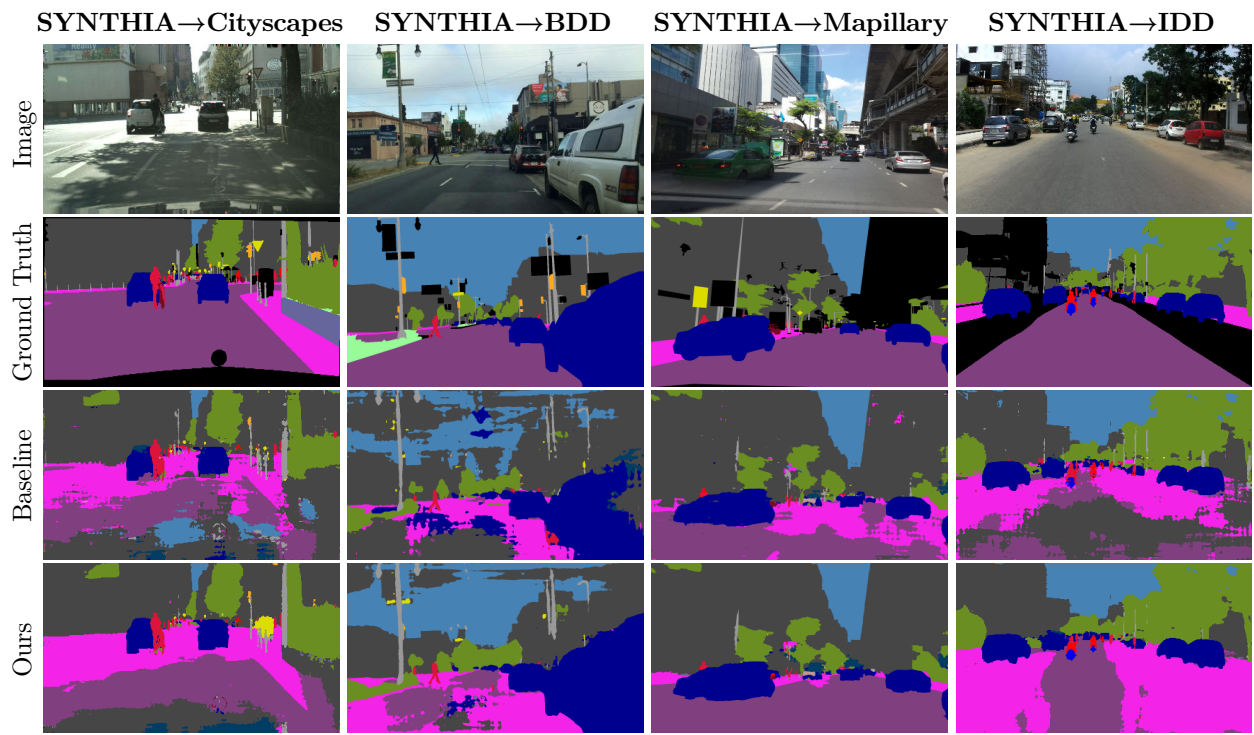
\begin{figure*}[!t]%
    \def\svgwidth{1.0\linewidth}
    \input{supp/figures/fresults_fig_syn50.tex}
    \caption{\textit{Qualitative semantic segmentation results for generalization from \synthia to Cityscapes, BDD, Mapillary, and IDD} for the ResNet-50 backbone.}
    \label{fig:qual_res_syn50}
 \end{figure*}

We provide additional qualitative results by running inference on three models: ResNet-101 trained on \gta in \cref{fig:qual_res_gta101}; ResNet-50 and ResNet-101 trained on \synthia in \cref{fig:qual_res_syn50,fig:qual_res_syn101}, respectively.
Similar to our observations in \cref{sec:exp_iabn_ttt} (\cf main text), our approach exhibits more homogeneous semantic masks with visibly fewer jagged-shaped artifacts than the baseline (\eg, ``sidewalk'' false positives in \cref{fig:qual_res_gta101}).
Models with self-adaptation may still struggle with cases of mislabeling regions with an incorrect, but semantically related class.
For example, the model often assigns ``sidewalk'' to the road pixels from BDD and IDD in \cref{fig:qual_res_syn50,fig:qual_res_syn101}.
This is an expected outcome if the erroneous labels are already contained in the pseudo-labels of the initial prediction, on which self-adaptation relies.
These failure cases occur more frequently if the domain shift between the train and the test distributions is more significant, such as between \synthia and \bdd, which can lead to poorly calibrated predictions.
Since applying \iabn alone results in improved calibration (\cf \cref{table:calibration}(b)), it alleviates this issue, and hence self-adaptation can cope well with milder domain shift scenarios, as a result.
As the examples on Cityscapes and Mapillary in \cref{fig:qual_res_syn50,fig:qual_res_syn101} show, despite the baseline model exhibiting some degree of this failure mode, our inference method visibly rectifies these errors.

We additionally ran our inference on video sequences and include the results as part of our supplementary material.
Confirming our previous analysis of the qualitative results (\cf \cref{sec:exp_ttt}), we observe that our approach clearly improves the segmentation quality and removes some of the most pathological failure modes of the baseline (\eg, the lower middle part of the frame area).

 \begin{figure*}[!t]%
    \def\svgwidth{1.0\linewidth}
    \input{supp/figures/fresults_fig_syn101.tex}
    \caption{\textit{Qualitative semantic segmentation results for generalization from \synthia to Cityscapes, BDD, Mapillary, and IDD} for the ResNet-101 backbone.}
    \label{fig:qual_res_syn101}
 \end{figure*}
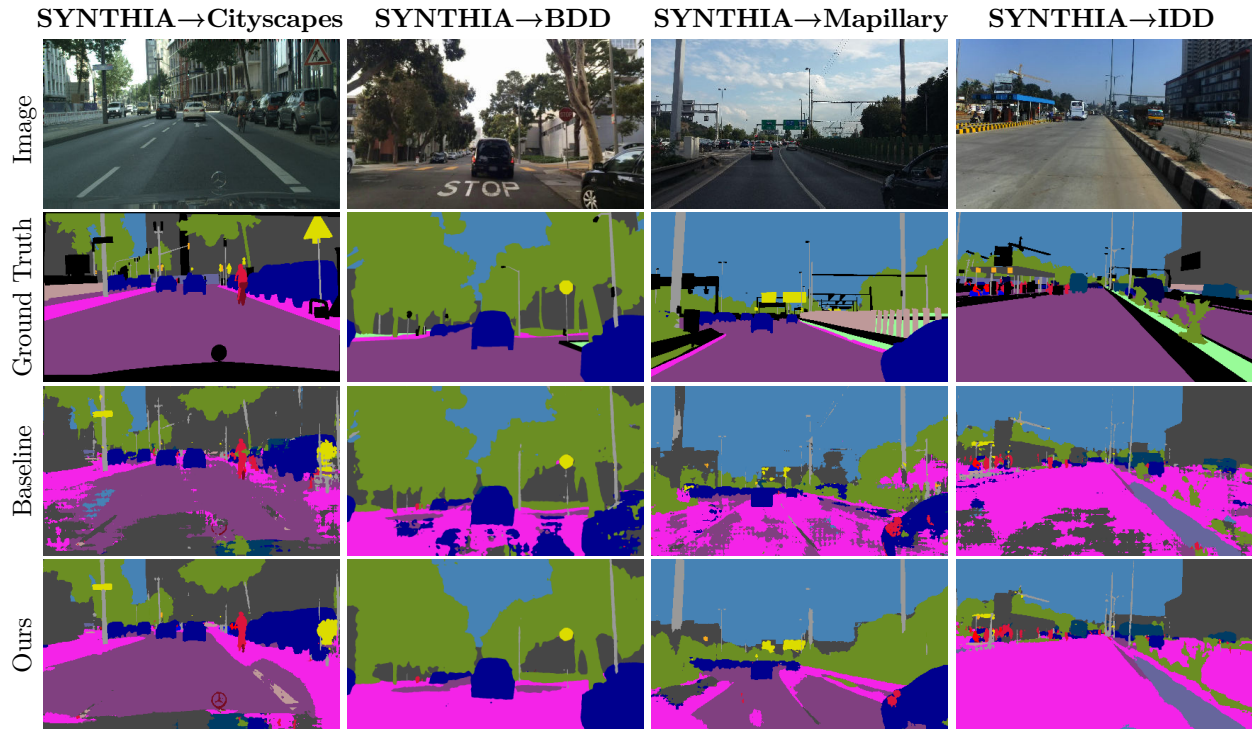
 
 \begin{figure*}[!t]%
    \def\svgwidth{1.0\linewidth}
    \input{supp/figures/aux/rn50gta_failure_examples.tex}
    \caption{\textit{Failure cases of semantic segmentation for generalization results from \gta to Cityscapes, BDD, Mapillary, and IDD} for the ResNet-50 backbone.}
    \label{fig:qual_failure_res_gta50}
 \end{figure*}
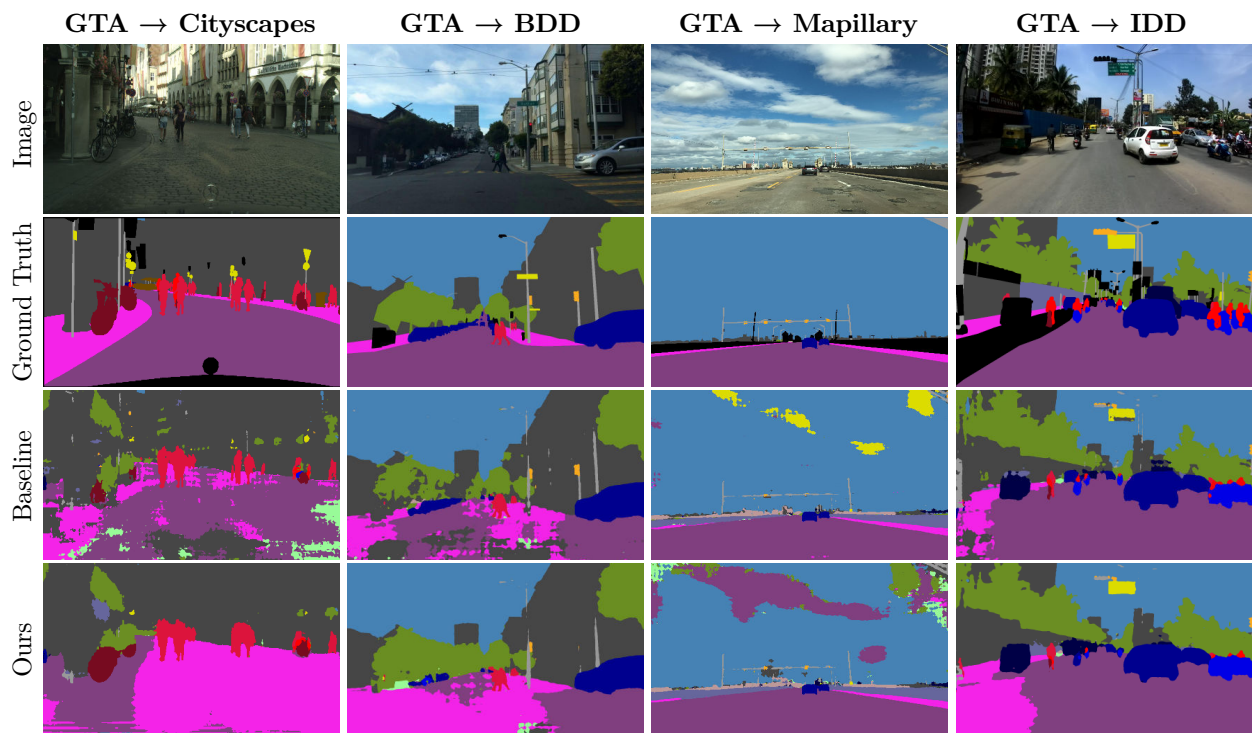

%% file: supp/figures/fresults_fig_gta101.tex
\begingroup%
  \makeatletter%
  \providecommand\color[2][]{%
    \errmessage{(Inkscape) Color is used for the text in Inkscape, but the package 'color.sty' is not loaded}%
    \renewcommand\color[2][]{}%
  }%
  \providecommand\transparent[1]{%
    \errmessage{(Inkscape) Transparency is used (non-zero) for the text in Inkscape, but the package 'transparent.sty' is not loaded}%
    \renewcommand\transparent[1]{}%
  }%
  \providecommand\rotatebox[2]{#2}%
  \newcommand*\fsize{\dimexpr\f@size pt\relax}%
  \newcommand*\lineheight[1]{\fontsize{\fsize}{#1\fsize}\selectfont}%
  \ifx\svgwidth\undefined%
    \setlength{\unitlength}{416.44803692bp}%
    \ifx\svgscale\undefined%
      \relax%
    \else%
      \setlength{\unitlength}{\unitlength * \real{\svgscale}}%
    \fi%
  \else%
    \setlength{\unitlength}{\svgwidth}%
  \fi%
  \global\let\svgwidth\undefined%
  \global\let\svgscale\undefined%
  \makeatother%
  \begin{picture}(1, 0.58789823)%
  \lsstyle
    \lineheight{1}%
    \setlength\tabcolsep{0pt}%
    \put(0,0){\includegraphics[width=\unitlength,page=1]{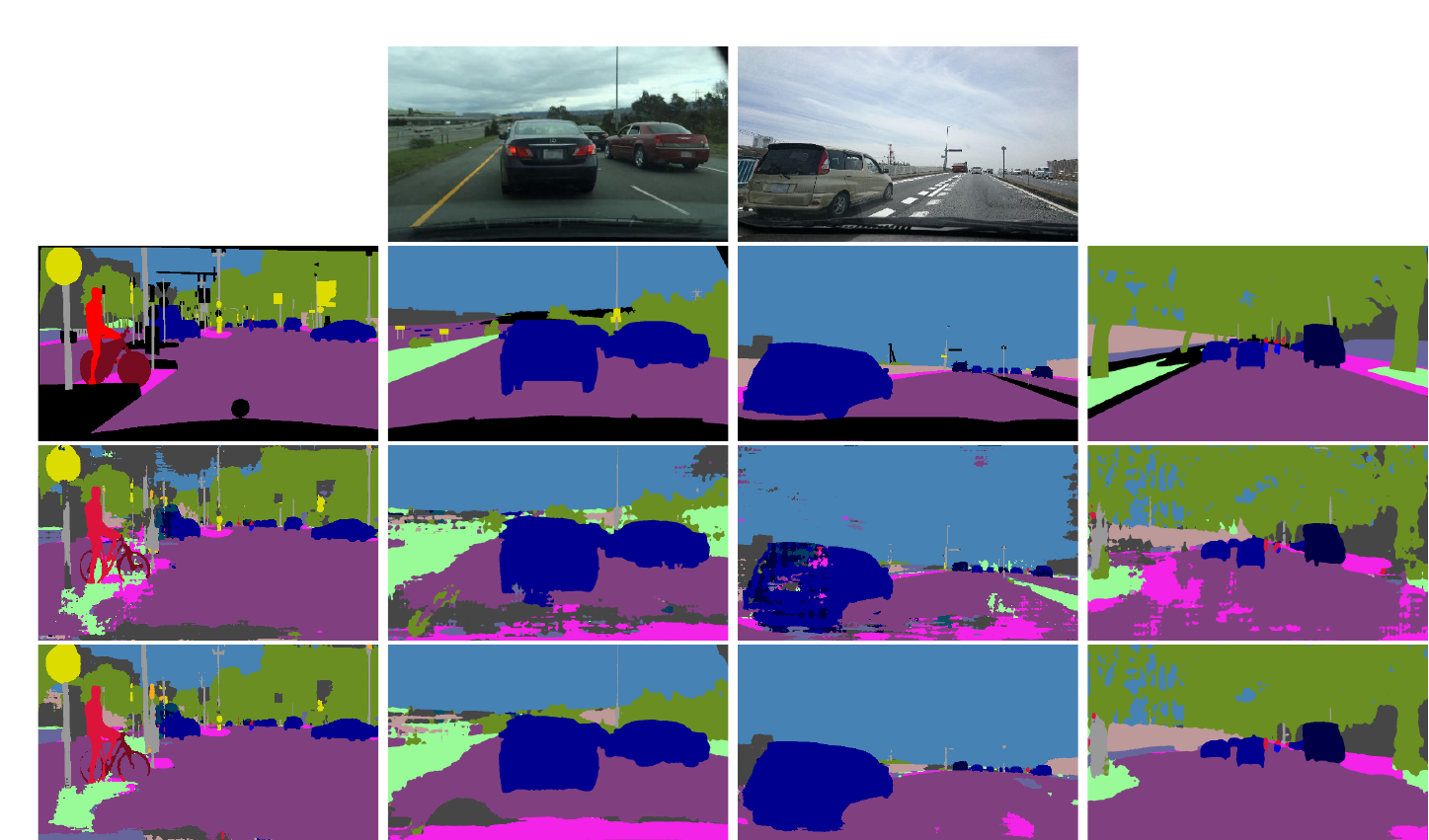}}%
    \put(0,0){\includegraphics[width=\unitlength,page=2]{supp/figures/fresults_fig_gta101_im.pdf}}%
    \put(0.145,0.565){\makebox(0,0)[ct]{\lineheight{1.25}\smash{\begin{tabular}[t]{l}\textbf{GTA → Cityscapes} \end{tabular}}}}%
    \put(0.387,0.565){\makebox(0,0)[ct]{\lineheight{1.25}\smash{\begin{tabular}[t]{l}\textbf{GTA → BDD} \end{tabular}}}}%
    \put(0.635,0.565){\makebox(0,0)[ct]{\lineheight{1.25}\smash{\begin{tabular}[t]{l}\textbf{GTA → Mapillary} \end{tabular}}}}%
    \put(0.879,0.565){\makebox(0,0)[ct]{\lineheight{1.25}\smash{\begin{tabular}[t]{l}\textbf{GTA → IDD} \end{tabular}}}}%

    \put(0.017,0.455){\rotatebox{90}{\makebox(0,0)[lt]{\lineheight{1.25}\smash{\begin{tabular}[t]{l}Image \end{tabular}}}}}%
    \put(0.017,0.283){\rotatebox{90}{\makebox(0,0)[lt]{\lineheight{1.25}\smash{\begin{tabular}[t]{l}Ground Truth\end{tabular}}}}}%
    \put(0.017,0.17){\rotatebox{90}{\makebox(0,0)[lt]{\lineheight{1.25}\smash{\begin{tabular}[t]{l}Baseline \end{tabular}}}}}%
    \put(0.017,0.045){\rotatebox{90}{\makebox(0,0)[lt]{\lineheight{1.25}\smash{\begin{tabular}[t]{l}Ours\end{tabular}}}}}%
  \end{picture}%
\endgroup%

%% file: supp/figures/fresults_fig_syn50.tex
\begingroup%
  \makeatletter%
  \providecommand\color[2][]{%
    \errmessage{(Inkscape) Color is used for the text in Inkscape, but the package 'color.sty' is not loaded}%
    \renewcommand\color[2][]{}%
  }%
  \providecommand\transparent[1]{%
    \errmessage{(Inkscape) Transparency is used (non-zero) for the text in Inkscape, but the package 'transparent.sty' is not loaded}%
    \renewcommand\transparent[1]{}%
  }%
  \providecommand\rotatebox[2]{#2}%
  \newcommand*\fsize{\dimexpr\f@size pt\relax}%
  \newcommand*\lineheight[1]{\fontsize{\fsize}{#1\fsize}\selectfont}%
  \ifx\svgwidth\undefined%
    \setlength{\unitlength}{416.44803692bp}%
    \ifx\svgscale\undefined%
      \relax%
    \else%
      \setlength{\unitlength}{\unitlength * \real{\svgscale}}%
    \fi%
  \else%
    \setlength{\unitlength}{\svgwidth}%
  \fi%
  \global\let\svgwidth\undefined%
  \global\let\svgscale\undefined%
  \makeatother%
  \begin{picture}(1, 0.58789823)%
  \lsstyle
    \lineheight{1}%
    \setlength\tabcolsep{0pt}%
    \put(0,0){\includegraphics[width=\unitlength,page=1]{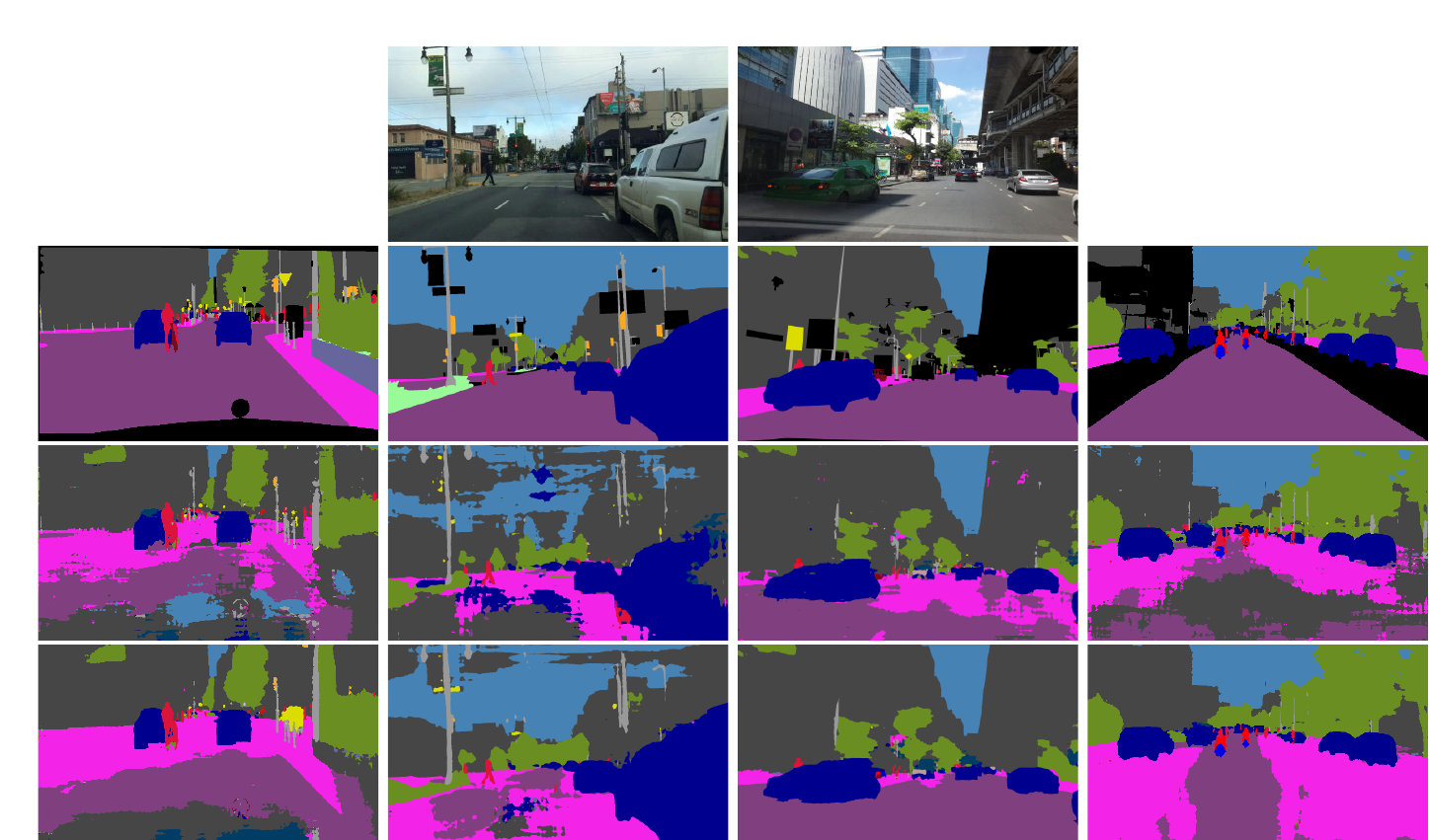}}%
    \put(0,0){\includegraphics[width=\unitlength,page=2]{supp/figures/fresults_fig_syn50_im.pdf}}%
    \put(0.145,0.565){\makebox(0,0)[ct]{\lineheight{1.25}\smash{\begin{tabular}[t]{l}\textbf{SYNTHIA→Cityscapes} \end{tabular}}}}%
    \put(0.389,0.565){\makebox(0,0)[ct]{\lineheight{1.25}\smash{\begin{tabular}[t]{l}\textbf{SYNTHIA→BDD} \end{tabular}}}}%
    \put(0.637,0.565){\makebox(0,0)[ct]{\lineheight{1.25}\smash{\begin{tabular}[t]{l}\textbf{SYNTHIA→Mapillary} \end{tabular}}}}%
    \put(0.879,0.565){\makebox(0,0)[ct]{\lineheight{1.25}\smash{\begin{tabular}[t]{l}\textbf{SYNTHIA→IDD} \end{tabular}}}}%

    \put(0.017,0.455){\rotatebox{90}{\makebox(0,0)[lt]{\lineheight{1.25}\smash{\begin{tabular}[t]{l}Image \end{tabular}}}}}%
    \put(0.017,0.283){\rotatebox{90}{\makebox(0,0)[lt]{\lineheight{1.25}\smash{\begin{tabular}[t]{l}Ground Truth\end{tabular}}}}}%
    \put(0.017,0.17){\rotatebox{90}{\makebox(0,0)[lt]{\lineheight{1.25}\smash{\begin{tabular}[t]{l}Baseline \end{tabular}}}}}%
    \put(0.017,0.045){\rotatebox{90}{\makebox(0,0)[lt]{\lineheight{1.25}\smash{\begin{tabular}[t]{l}Ours\end{tabular}}}}}%
  \end{picture}%
\endgroup%

%% file: supp/figures/fresults_fig_syn101.tex
\begingroup%
  \makeatletter%
  \providecommand\color[2][]{%
    \errmessage{(Inkscape) Color is used for the text in Inkscape, but the package 'color.sty' is not loaded}%
    \renewcommand\color[2][]{}%
  }%
  \providecommand\transparent[1]{%
    \errmessage{(Inkscape) Transparency is used (non-zero) for the text in Inkscape, but the package 'transparent.sty' is not loaded}%
    \renewcommand\transparent[1]{}%
  }%
  \providecommand\rotatebox[2]{#2}%
  \newcommand*\fsize{\dimexpr\f@size pt\relax}%
  \newcommand*\lineheight[1]{\fontsize{\fsize}{#1\fsize}\selectfont}%
  \ifx\svgwidth\undefined%
    \setlength{\unitlength}{416.44803692bp}%
    \ifx\svgscale\undefined%
      \relax%
    \else%
      \setlength{\unitlength}{\unitlength * \real{\svgscale}}%
    \fi%
  \else%
    \setlength{\unitlength}{\svgwidth}%
  \fi%
  \global\let\svgwidth\undefined%
  \global\let\svgscale\undefined%
  \makeatother%
  \begin{picture}(1, 0.58789823)%
  \lsstyle
    \lineheight{1}%
    \setlength\tabcolsep{0pt}%
    \put(0,0){\includegraphics[width=\unitlength,page=1]{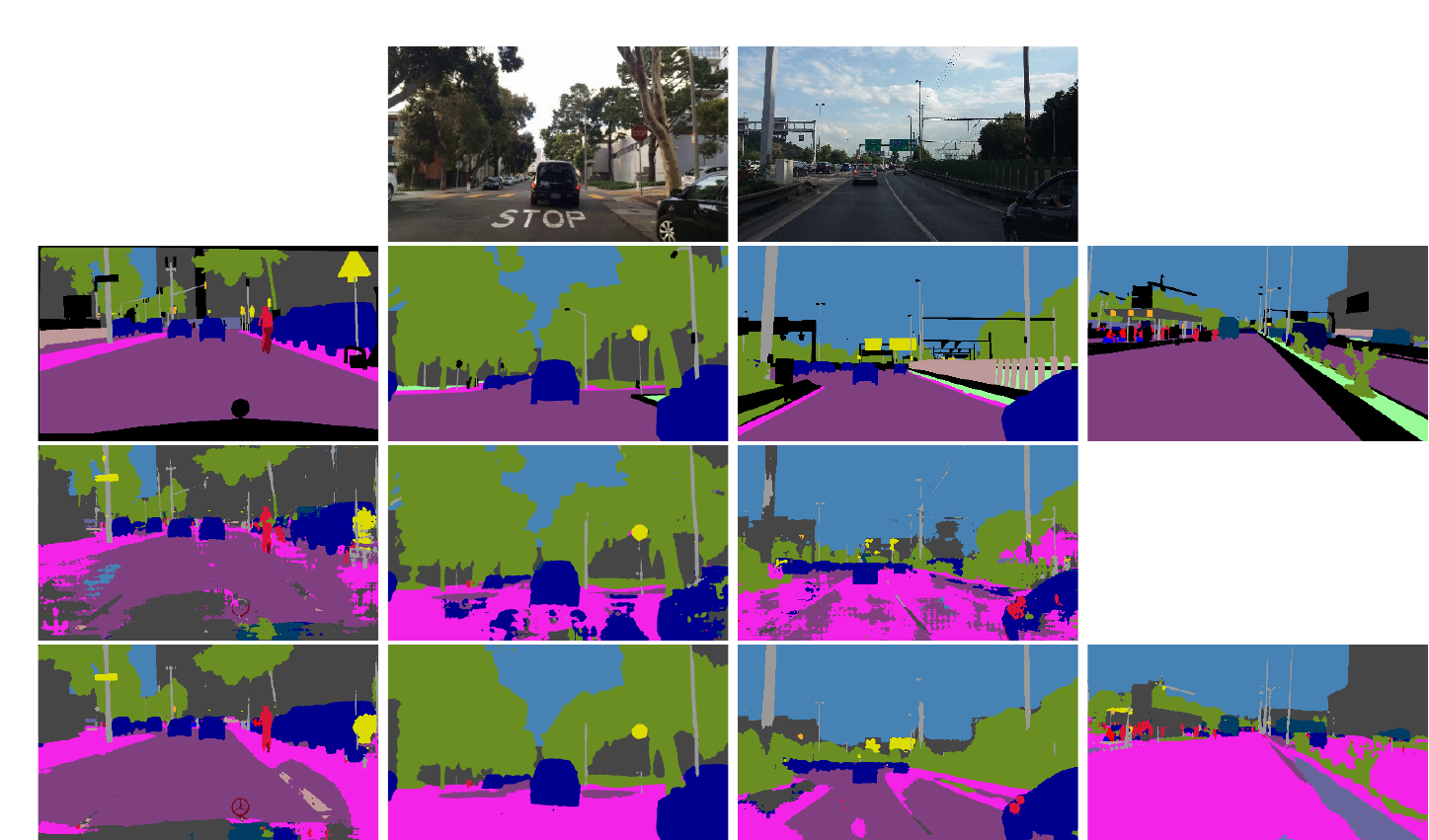}}%
    \put(0,0){\includegraphics[width=\unitlength,page=2]{supp/figures/fresults_fig_syn101_im.pdf}}%
    \put(0.145,0.565){\makebox(0,0)[ct]{\lineheight{1.25}\smash{\begin{tabular}[t]{l}\textbf{SYNTHIA→Cityscapes} \end{tabular}}}}%
    \put(0.389,0.565){\makebox(0,0)[ct]{\lineheight{1.25}\smash{\begin{tabular}[t]{l}\textbf{SYNTHIA→BDD} \end{tabular}}}}%
    \put(0.637,0.565){\makebox(0,0)[ct]{\lineheight{1.25}\smash{\begin{tabular}[t]{l}\textbf{SYNTHIA→Mapillary} \end{tabular}}}}%
    \put(0.879,0.565){\makebox(0,0)[ct]{\lineheight{1.25}\smash{\begin{tabular}[t]{l}\textbf{SYNTHIA→IDD} \end{tabular}}}}%

    \put(0.017,0.455){\rotatebox{90}{\makebox(0,0)[lt]{\lineheight{1.25}\smash{\begin{tabular}[t]{l}Image \end{tabular}}}}}%
    \put(0.017,0.283){\rotatebox{90}{\makebox(0,0)[lt]{\lineheight{1.25}\smash{\begin{tabular}[t]{l}Ground Truth\end{tabular}}}}}%
    \put(0.017,0.17){\rotatebox{90}{\makebox(0,0)[lt]{\lineheight{1.25}\smash{\begin{tabular}[t]{l}Baseline \end{tabular}}}}}%
    \put(0.017,0.045){\rotatebox{90}{\makebox(0,0)[lt]{\lineheight{1.25}\smash{\begin{tabular}[t]{l}Ours\end{tabular}}}}}%
  \end{picture}%
\endgroup%

%% file: supp/figures/aux/rn50gta_failure_examples.tex
\begingroup%
  \makeatletter%
  \providecommand\color[2][]{%
    \errmessage{(Inkscape) Color is used for the text in Inkscape, but the package 'color.sty' is not loaded}%
    \renewcommand\color[2][]{}%
  }%
  \providecommand\transparent[1]{%
    \errmessage{(Inkscape) Transparency is used (non-zero) for the text in Inkscape, but the package 'transparent.sty' is not loaded}%
    \renewcommand\transparent[1]{}%
  }%
  \providecommand\rotatebox[2]{#2}%
  \newcommand*\fsize{\dimexpr\f@size pt\relax}%
  \newcommand*\lineheight[1]{\fontsize{\fsize}{#1\fsize}\selectfont}%
  \ifx\svgwidth\undefined%
    \setlength{\unitlength}{416.44803692bp}%
    \ifx\svgscale\undefined%
      \relax%
    \else%
      \setlength{\unitlength}{\unitlength * \real{\svgscale}}%
    \fi%
  \else%
    \setlength{\unitlength}{\svgwidth}%
  \fi%
  \global\let\svgwidth\undefined%
  \global\let\svgscale\undefined%
  \makeatother%
  \begin{picture}(1, 0.58789823)%
  \lsstyle
    \lineheight{1}%
    \setlength\tabcolsep{0pt}%
    \put(0,0){\includegraphics[width=\unitlength,page=1]{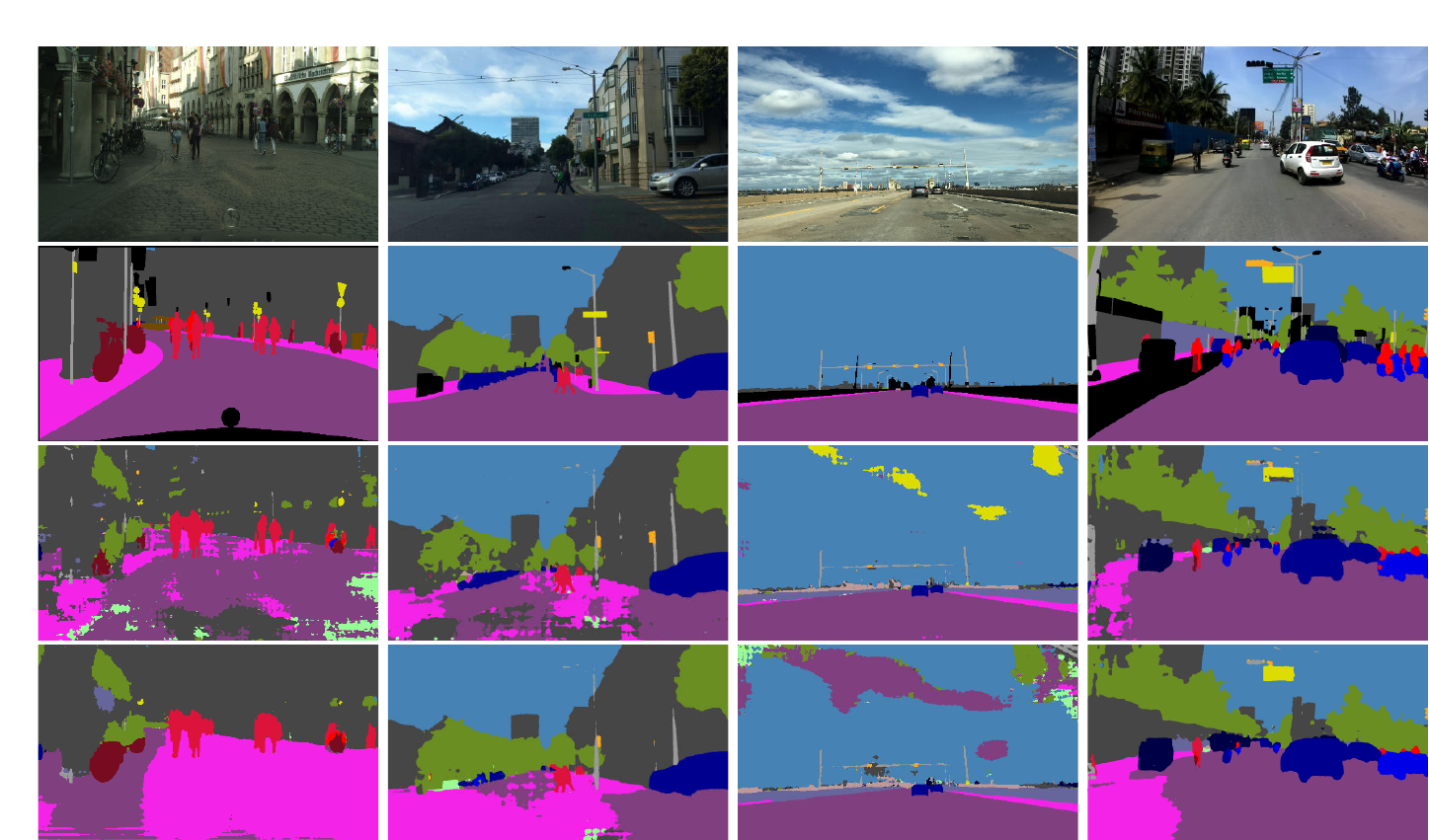}}%
    \put(0.145,0.565){\makebox(0,0)[ct]{\lineheight{1.25}\smash{\begin{tabular}[t]{l}\textbf{GTA → Cityscapes} \end{tabular}}}}%
    \put(0.387,0.565){\makebox(0,0)[ct]{\lineheight{1.25}\smash{\begin{tabular}[t]{l}\textbf{GTA → BDD} \end{tabular}}}}%
    \put(0.635,0.565){\makebox(0,0)[ct]{\lineheight{1.25}\smash{\begin{tabular}[t]{l}\textbf{GTA → Mapillary} \end{tabular}}}}%
    \put(0.879,0.565){\makebox(0,0)[ct]{\lineheight{1.25}\smash{\begin{tabular}[t]{l}\textbf{GTA → IDD} \end{tabular}}}}%

    \put(0.017,0.455){\rotatebox{90}{\makebox(0,0)[lt]{\lineheight{1.25}\smash{\begin{tabular}[t]{l}Image \end{tabular}}}}}%
    \put(0.017,0.283){\rotatebox{90}{\makebox(0,0)[lt]{\lineheight{1.25}\smash{\begin{tabular}[t]{l}Ground Truth\end{tabular}}}}}%
    \put(0.017,0.17){\rotatebox{90}{\makebox(0,0)[lt]{\lineheight{1.25}\smash{\begin{tabular}[t]{l}Baseline \end{tabular}}}}}%
    \put(0.017,0.045){\rotatebox{90}{\makebox(0,0)[lt]{\lineheight{1.25}\smash{\begin{tabular}[t]{l}Ours\end{tabular}}}}}%
  \end{picture}%
\endgroup%

%% file: supp/sections/supp_failure.tex
\begin{figure*}[ht!]%
    \includegraphics[width=1.0\linewidth]{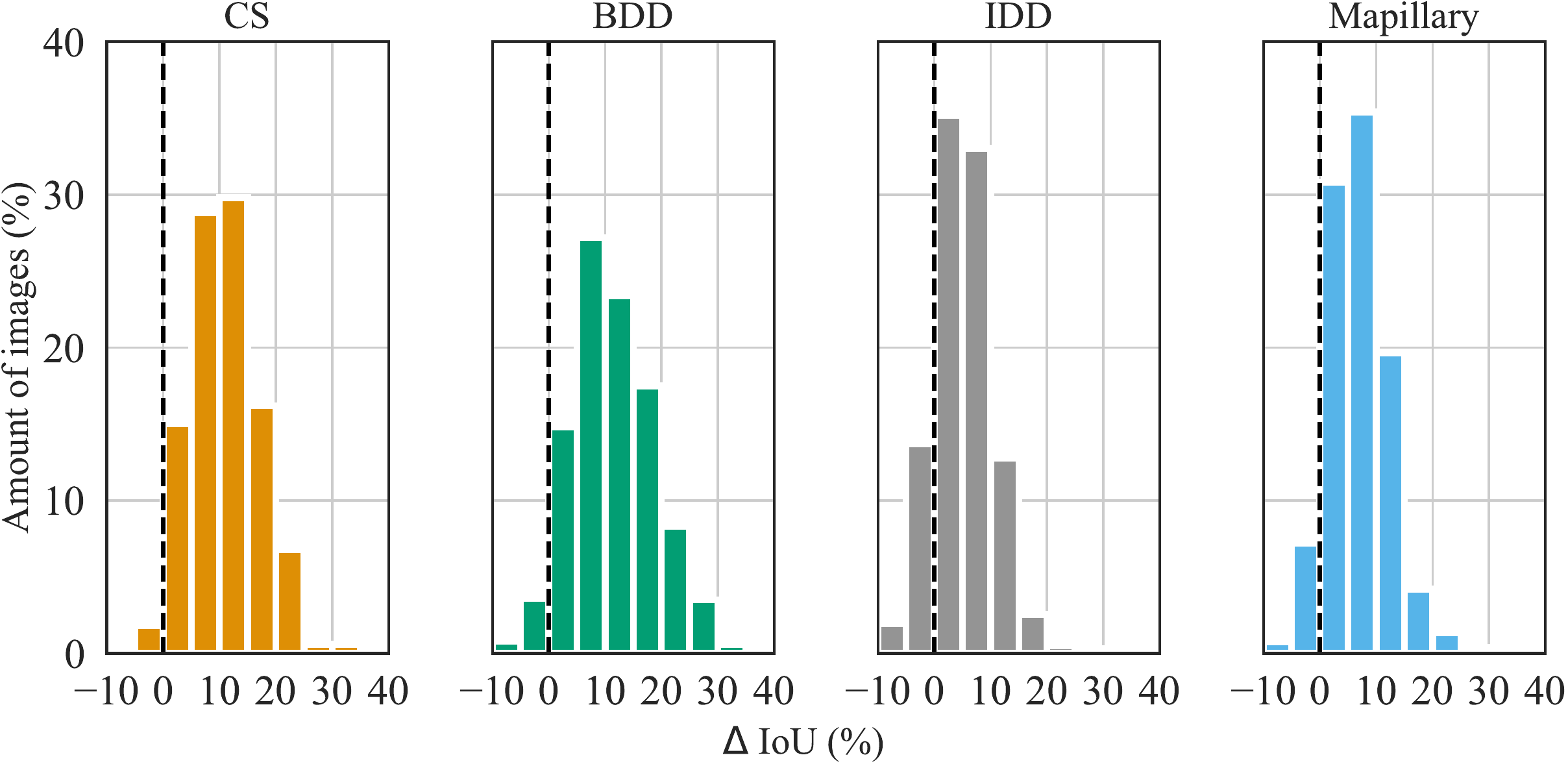}
    \caption{\textit{Empirical distribution of the IoU change of individual images for generalization from source domain (\gta) to target domains (Cityscapes, BDD, IDD and Mapillary)} for the ResNet-50 backbone. We visualize the relative improvement of our self-adaptive inference strategy \wrt\ the baseline in terms of accuracy with $\Delta$ IoU (\%).}
    \label{fig:density_failure_res_gta50}
 \end{figure*}

Conceptually, if the initial semantic prediction is incorrect and confident, which may happen due to imperfect model calibration, such error is likely to end up in the pseudo-labels and lead astray the self-adaptation process.
The rightmost columns in \cref{fig:qual_res_syn50,fig:qual_res_syn101} already visualized this failure mode: the baseline assigns large areas of the ``road'' class to the  ``sidewalk'' category.
We have further extended these examples with \cref{fig:qual_failure_res_gta50}.
Misguided by incorrect pseudo-labels, our self-adaptation may exacerbate this issue, \ie, propagate the incorrect label to the areas sharing the same appearance.
To gain further insights, we investigated this issue statistically.
The histograms in \cref{fig:density_failure_res_gta50} show relative improvement of our self-adaptive inference strategy \wrt\ the baseline in terms of IoU on four validation sets.
We conclude that such decrease in segmentation accuracy is actually quite rare: less than 10\% of the images, on average, exhibit lower accuracy.
In most such cases the accuracy reduces only marginally (by less than 5\%), and only a fraction of images -- less than 1\% on average -- deteriorate in accuracy by at most 10\%.
The overwhelming majority of the image samples benefit from our self-adaptive process, which increases their  accuracy by up to 35\% IoU compared to the baseline.

%% file: supp/sections/supp_dataset.tex
\paragraph{\gta.}
GTA \citep{Richter:2016:TSS} is a street view dataset generated semi-automatically from the computer game Grand Theft Auto V.
The dataset consists of 12,403 training images, 6,382 validation images, and 6,181 testing images of resolution 1914 × 1052 with 19 different semantic classes.

\myparagraph{\synthia.}
We use the SYNTHIA-RAND-CITYSCAPES subset of the synthetic dataset SYNTHIA \citep{Ros:2016:SDL}, which contains 9,400 images, and has 16 semantic classes in common with GTA.
Images have a resolution of 1280 × 760 pixels.

\myparagraph{\wilddash.}
The \wilddash benchmark \citep{ZendelHMSD18} was developed to evaluate models regarding their robustness for driving scenarios under real-world conditions.
It comprises 4,256 images of real-world scenes with a resolution of 1920 × 1080 pixels.

\myparagraph{Cityscapes.}
Cityscapes \citep{Cordts:2016:CDS} is an ego-centric street-scene dataset and contains 5,000 high-resolution images with 2048 × 1024 pixels.
It is split into 2,975 train, 500 val, and 1,525 test images with 19 semantic classes being annotated.

\myparagraph{BDD.}
BDD \citep{Yu:2020:BDD} is a driving video dataset, which also contains semantic labelings with the identical 19 classes as in the other datasets.
Images have a resolution of 1280 × 720 pixels.
The training, validation, and test sets contain 7,000, 1,000, and 2,000 images, respectively.

\myparagraph{IDD.}
IDD \citep{Varma:2019:IDD} is a dataset for road scene understanding in unstructured environments.
It contains 10,003 images annotated with 34 classes even though we only evaluate on the 19 classes overlapping with the other datasets.
IDD is split into 6,993 training images, 981 validation images, and 2,029 test images.

\myparagraph{Mapillary.}
Annotations from Mapillary \citep{Neuhold:2017:MVD} contain 66 object classes; analogously to IDD we only evaluate on the 19 classes overlapping with the other datasets.
The dataset is split into a training set with 18,000 images and a validation set with 2,000 images with a minimum resolution of 1920 × 1080 pixels.